\documentclass[]{gtech}

\usepackage{amssymb}
\usepackage{multirow}
\usepackage{bigdelim}
\usepackage{todonotes}
\usepackage{longtable}
\usepackage{tabularray}
\usepackage{wrapfig}
\usepackage[most]{tcolorbox} 
\usepackage{xcolor}
\usepackage{url}
\usepackage{algorithm}  
\usepackage{algpseudocode} 
\usepackage[toc,page]{appendix}
\usepackage{multirow}
\usepackage[normalem]{ulem}
\usepackage{adjustbox}
\usepackage{booktabs}  
\usepackage{natbib}
\usepackage{colortbl}  
\usepackage{subcaption}
\usepackage{pifont}
\usepackage{amsmath}
\usepackage{amsfonts}       
\usepackage{multirow}
\usepackage{mathrsfs}
\usepackage{tabularx}
\usepackage[utf8]{inputenc} 
\usepackage[T1]{fontenc}    
\usepackage{hyperref}       
\usepackage{url}            
\usepackage{booktabs}       
\usepackage{nicefrac}       
\usepackage{microtype}      
\usepackage{xcolor}         
\definecolor{ourreward}{HTML}{32bbee}
\definecolor{sparsereward}{HTML}{ed9a81}
\usepackage{graphicx}
\usepackage{wrapfig}
\usepackage{cleveref}
\usepackage{fvextra}
\tcbuselibrary{listings, breakable}

\usepackage{bbding}
\usepackage{perpage}
\MakePerPage{footnote}
\usepackage{caption}
\usepackage{CJK}

\usepackage{bbm}
\useunder{\uline}{\ul}{}

\newcommand{\ourmodel}{Skill-Venus}
\newcommand{\retrievalbenchmark}{SkillBench-Venus-S}
\newcommand{\retrievalbenchmarkfull}{\textbf{\retrievalbenchmark}earch}
\newcommand{\taskbenchmark}{SkillBench-Venus-E}
\newcommand{\taskbenchmarkfull}{\textbf{\taskbenchmark}nd2end}

\title{Skill-Venus: Towards Agent-Oriented Skill Search}

\author{Venus Team, Ant Group}

\abstract{
\vspace{-1em}
\begin{center}
\bfseries Abstract
\end{center}
\vspace{-0.5em}
As large language model (LLM) agents increasingly rely on external skills to solve complex tasks, effectively retrieving useful skills from large repositories has become essential. However, existing skill search methods primarily rely on the initial user request, overlooking the evolving capability needs that arise as agents execute tasks. To address this limitation, we introduce \textbf{Skill-Venus}, an agent-oriented skill search framework that enables a new paradigm of on-demand skill acquisition through a skill-search meta-skill. Agents use the meta-skill to autonomously formulate queries for the current step, inspect candidate skills, load suitable ones, and search again as new needs arise. 
Training a skill search model for this paradigm requires data that capture both what agents search for during execution and which skills help them make progress.
To obtain these data, we develop a skill-grounded synthesis pipeline that produces nearly 100K executable tasks and collect trajectories from agents performing them.
From these trajectories, we extract search queries and derive supervision based on skill relevance, adoption, and utility to train our skill searcher with a masked contrastive objective. We further construct complementary benchmarks for offline skill search and end-to-end task execution. Extensive experiments demonstrate that Skill-Venus improves skill search performance over corresponding base models. Notably, across three agent models, Skill-Venus improves average end-to-end task scores by 6.3\% while reducing skill search calls by 26.4\%, demonstrating the value of skill search aligned with agents’ evolving capability needs.
}

\gtechdata[Code \& Benchmark]{\url{https://github.com/inclusionAI/Skill-Venus}}
\gtechdata[Model \& Data]
{\url{https://huggingface.co/collections/inclusionAI/skill-venus}}

\begin{document}
\maketitle

\begin{figure}[ht]
    \centering
    \vspace{1em}
    \includegraphics[width=\linewidth]{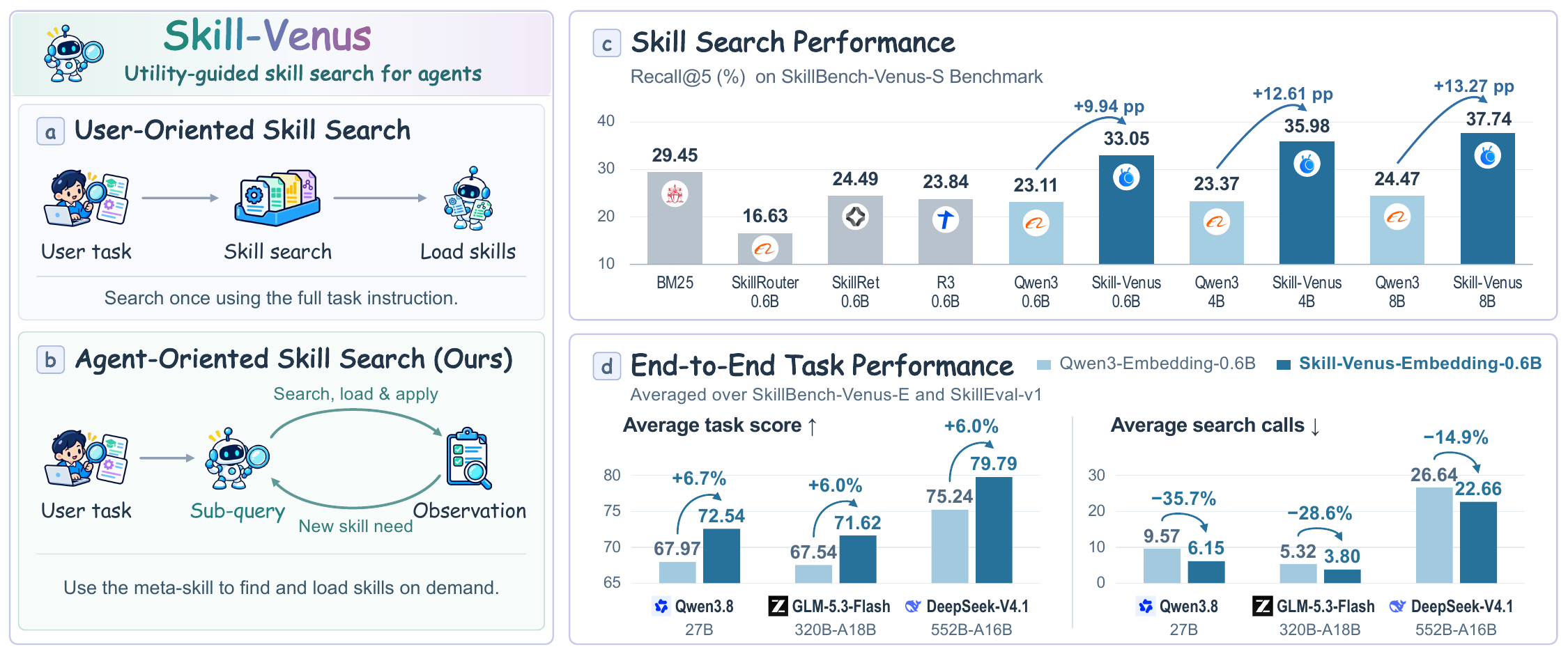}
    \caption{
    \textbf{Motivation and results of Skill-Venus.}
    (a) User-oriented skill search.
    (b) Agent-oriented skill search.
    (c) Skill search performance on SkillBench-Venus-S.
    (d) End-to-end task scores and search calls across three agent models using 0.6B models.
    }
    \label{fig:hero_abstract}
\end{figure}

\newpage
\section{Introduction}
\begin{figure}[h]
    \centering
    \vspace{1em}
    \includegraphics[width=\linewidth]{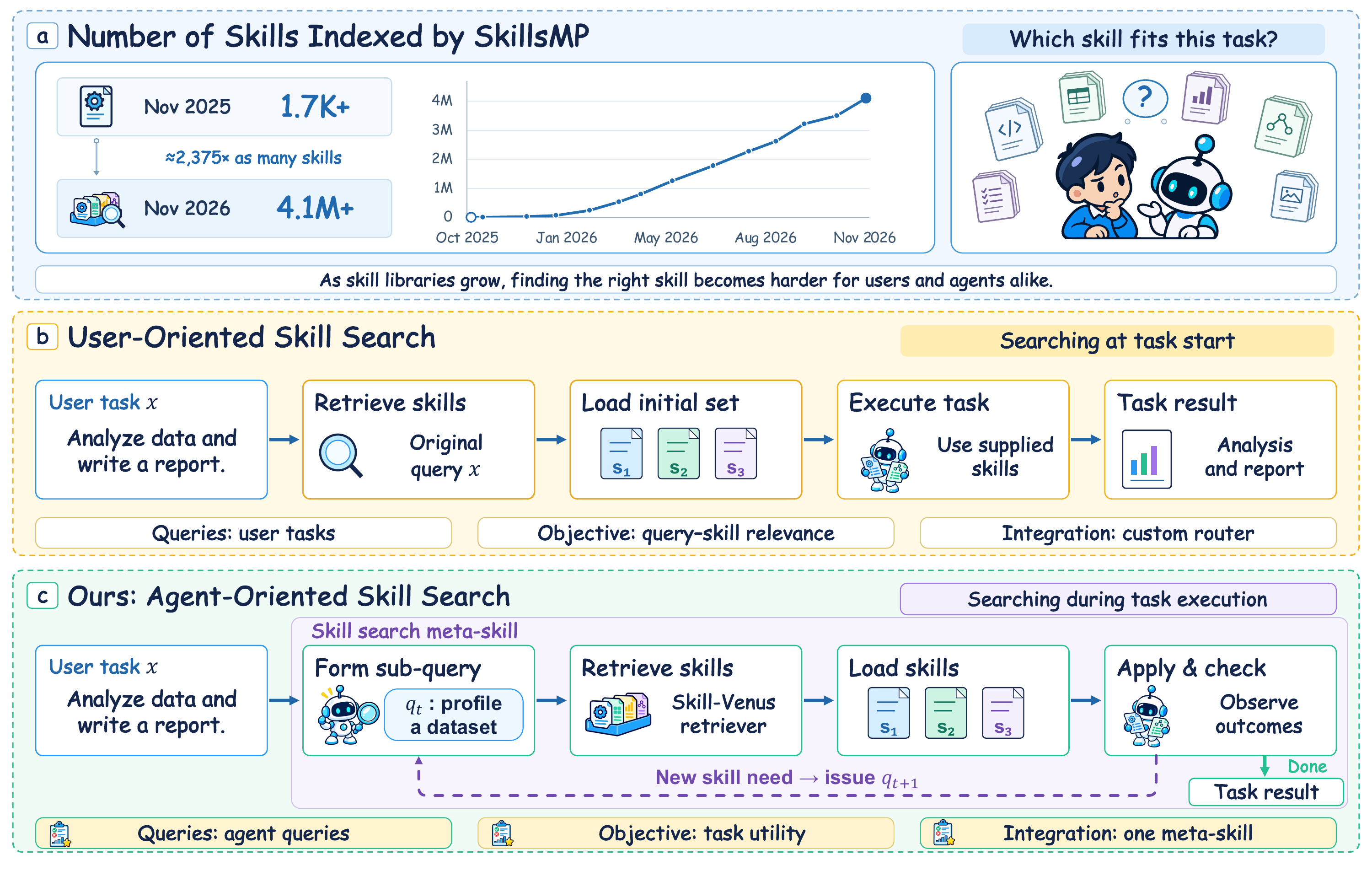}
    \caption{Growth of skill repositories and comparison of skill search strategies. (a) SkillsMP grew from 1,727 indexed skills to over 4.1 million in about a year. (b) User-oriented search retrieves skills once, using the full user task as the query. (c) Agent-oriented search dynamically retrieves skills during task execution based on intermediate capability needs and task utility.}
    \label{fig:hero_intro}
\end{figure}

Agent skills provide reusable procedural instructions and resources that extend the capabilities of large language model (LLM) agents. By equipping agents with specialized workflows, skills enable them to handle tasks that go beyond their intrinsic capabilities. However, as the ecosystem of agent skills expands, finding the right skills from increasingly large repositories becomes a growing challenge. 
As shown in Figure~\ref{fig:hero_intro}(a), the number of skills indexed by SkillsMP\footnote{\url{https://skillsmp.com/}} has grown from 1,727 to over 4.1 million in approximately one year. Such rapid growth makes manual discovery increasingly impractical, while exposing all available skills to an agent would incur substantial context overhead. 
Although agent systems such as Codex and Claude Code adopt progressive disclosure to expose skill metadata and load full instructions on demand~\citep{openai2026skills,anthropic2026claudeskills}, this mechanism alone does not solve the problem of discovering relevant skills from a large external repository. 
Consequently, effective skill search is becoming essential for enabling agents to access specialized capabilities when needed~\citep{zheng2026skillrouter,kang2026skillret}.

Existing skill search methods primarily follow a user-oriented paradigm, as illustrated in Figure~\ref{fig:hero_intro}(b). They use the original user task as the skill search query and select a set of relevant skills before task execution begins~\citep{zheng2026skillrouter,kang2026skillret,wang2026r3}. This setup assumes that the skills required for a task can be identified from its initial instruction. However, complex tasks often involve multiple stages, and the agent's capability needs evolve as it interacts with tools, examines intermediate results, and encounters unexpected issues. A skill that appears relevant to the overall task may not address the specific operation the agent needs to perform at a particular stage. Conversely, a capability that was not apparent from the initial request may become necessary only after observing intermediate execution results. Retrieving skills only once at the beginning therefore limits the agent's ability to adapt its skill usage to evolving task requirements.

To address these limitations, we propose a skill-search meta-skill that shifts skill discovery from preselection based on the initial user request to on-demand acquisition during task execution, as illustrated in Figure~\ref{fig:hero_intro}(c). Under this agent-oriented skill search paradigm, agents formulate queries from their current execution states, search and load relevant skills, and search again as new needs arise. Realizing this paradigm poses two key challenges.
First, the skill search model should understand agent queries, which describe specific intermediate capability needs rather than the overall user task. Second, skill search should be guided not only by query--skill relevance but also by task utility~\citep{cheng2026skillutility, zhang2026beyond}: a skill is valuable when its instructions help the agent make progress toward completing the task. A skill may be topically relevant to a query yet provide procedures that are incompatible with the current environment or ineffective for the actual task. Therefore, learning an effective agent-oriented skill search model requires supervision that connects search queries and retrieved skills to their subsequent use and impact on task execution.

Developing such a skill search model, however, requires training data that capture both intermediate capability needs and the consequences of skill use. Existing large-scale skill search datasets primarily associate complete user tasks with query--skill  semantic relevance labels~\citep{kang2026skillret,su2026sra}. Although useful for conventional retrieval, these datasets provide limited coverage of queries generated during execution and lack the task environments and verifiers needed to assess the practical utility of retrieved skills. Meanwhile, existing benchmarks for evaluating skill use provide executable tasks but contain substantially fewer examples and are primarily designed for evaluation rather than large-scale training~\citep{li2026skillsbench,tan2026skt}. This motivates a unified data pipeline that constructs executable tasks from skills, collects agent execution trajectories, and derives skill search supervision from skill relevance, adoption, and observed progress.

To this end, we introduce \textbf{Skill-Venus}, an agent-oriented skill search framework. We first develop a skill-grounded data synthesis pipeline that constructs nearly 100,000 executable tasks from existing skills, each with inputs, an execution environment, and a task-specific verifier. We collect execution trajectories as agents perform these tasks using the skill-search meta-skill. By analyzing how agents retrieve and use skills throughout task execution, we derive query--skill supervision that captures not only whether a skill matches a query, but also whether using it helps the agent make progress toward completing the task. We use this supervision to train our skill searcher with a masked contrastive objective. To evaluate both skill search quality and downstream task performance, we construct two complementary benchmarks: an offline skill search benchmark with 1,000 agent queries and an end-to-end benchmark with 250 verified executable tasks.

Experiments show that Skill-Venus improves Recall@5 by up to 13.27 percentage points over corresponding base skill search models. More importantly, across three agent models, Skill-Venus improves average end-to-end task scores by 6.3\% while reducing skill search calls by 26.4\%, demonstrating that better skill search of agent-specific capabilities can translate into more effective task execution.
Our contributions are summarized as follows:

$\bullet$~\textbf{Paradigm.} We introduce \emph{agent-oriented skill search}, extending skill discovery beyond the initial user tasks. To support this paradigm, we package skill search as a meta-skill so agents can decide when to search, what to search for, and which skills to load as the task unfolds.

$\bullet$~\textbf{Method.} We develop a pipeline that synthesizes nearly 100K executable tasks from existing skills. From agent trajectories on these tasks, we extract 156K positive training pairs, each linking a search query to a skill the agent used to move the task forward.

$\bullet$~\textbf{Evaluation.} We introduce \retrievalbenchmark{} to evaluate skill ranking on 1,000 queries issued by agents during task execution and \taskbenchmark{} to measure how well agents complete 250 executable tasks using on-demand skill search, facilitating future research on this area.

$\bullet$~\textbf{Results.} Skill-Venus improves skill search performance over Qwen3-Embedding-0.6B on public benchmarks, achieves 6.3\% higher task scores with 26.4\% fewer search calls on average across three agents, and improves search quality on Ant ToolCenter, Ant Group's largest internal skill platform.

\section{Preliminary}
\label{sec:preliminaries}

\subsection{Skill Search as a Meta-Skill}
\label{sec:preliminaries:interaction}
An agent skill packages reusable procedural instructions, often together with resources such as scripts and templates, to support specialized tasks. As skill repositories grow, finding and preparing suitable skills places an increasing burden on users. Maintaining a large local collection is also costly, and additional skills may become necessary as a task progresses. These evolving needs call for a way to discover and acquire skills throughout task execution. Giving the agent this ability allows it to obtain suitable skills as they become necessary, reducing the user's effort in finding and preparing them.

To support this workflow, we package skill search and acquisition as a reusable skill, which we call a \textbf{skill-search meta-skill}. Its instructions guide the agent in deciding when to search, how to express its current capability need, and how to obtain and use suitable skills. Users can therefore submit tasks as usual, while the agent manages skill discovery as part of task execution.

When a capability need arises, the agent formulates a query and calls \textsc{Search}. It inspects the returned skill names and descriptions to identify suitable candidates, refining the query when necessary. Once a skill is selected, the agent calls \textsc{Fetch} to obtain its full instructions and supporting resources, then applies its guidance to the current task. Subsequent execution results may reveal further needs, prompting another search. This interaction connects skill discovery, acquisition, and use throughout task execution.

\subsection{Task Formulation}
\label{sec:preliminaries:task}

We represent an executable task as $\xi=(x,I,E,V)$, where $x$ is the user's task instruction, $I$ denotes the initial inputs, $E$ is the execution environment, and $V$ is the task verifier. Let $\mathcal{S}$ be the skill repository. In user-oriented skill search, the skill search model first searches for skills using the full task instruction $x$ as its query. The agent then starts execution with the skills returned by that search. In agent-oriented skill search, the agent uses a skill-search meta-skill to find and load skills as new needs arise during execution. Its initial context $h_0$ includes $x$, and it can access $I$ in $E$.
At turn $t$, the agent generates reasoning $r_t$ and takes an action $a_t$ based on the trajectory so far, $h_{t-1}$. If the agent calls a tool, it receives an observation $o_t$ and extends the trajectory to $h_t=(h_{t-1},r_t,a_t,o_t)$.

For a skill search action, $a_t=\textsc{Search}(q_t)$, the agent formulates $q_t$ to express its current capability need based on the task context and execution history. The skill search model $R_\theta$ ranks skills in $\mathcal{S}$ according to $q_t$ and returns an ordered list of the top $K$ candidates:
\begin{equation}
  \mathcal{C}_t = R_\theta(q_t;\mathcal{S})
  = (s_{t,1},\ldots,s_{t,K}).
  \label{eq:skill-retrieval}
\end{equation}
The agent inspects the candidates' names and descriptions and can call \textsc{Fetch} to obtain a selected skill's full instructions and supporting resources. Search and fetch results enter the execution history, informing subsequent actions and queries. The agent can then apply the skill, refine its search, or search for additional skills as the task progresses.

At the end of execution, the agent produces an outcome $y$, which may include a final response, generated artifacts, and changes to the environment. The verifier returns a task score $V(y)$ measuring how well this outcome satisfies the task requirements. We therefore should evaluate skill search through both the relevance of retrieved skills to agent queries and the downstream task performance achieved when agents use them.

\section{Methodology}
\label{sec:method}

To train a search model for agent-oriented skill search, we need to observe what agents search for during execution and whether the skills they find help them make progress. We therefore build a skill corpus and synthesize executable tasks in which the source skills can improve the outcome (Section~\ref{sec:method:data}). 
By running agents on these tasks, we collect trajectories from which we extract search queries and annotate the utility of retrieved skills (Section~\ref{sec:method:training}). We then use these utility-labeled query--skill pairs to train the skill search model (Section~\ref{sec:method:retriever-training}).

\subsection{Skill-Grounded Task Synthesis}
\label{sec:method:data}

\begin{figure}[t]
  \centering
  \includegraphics[width=\linewidth]{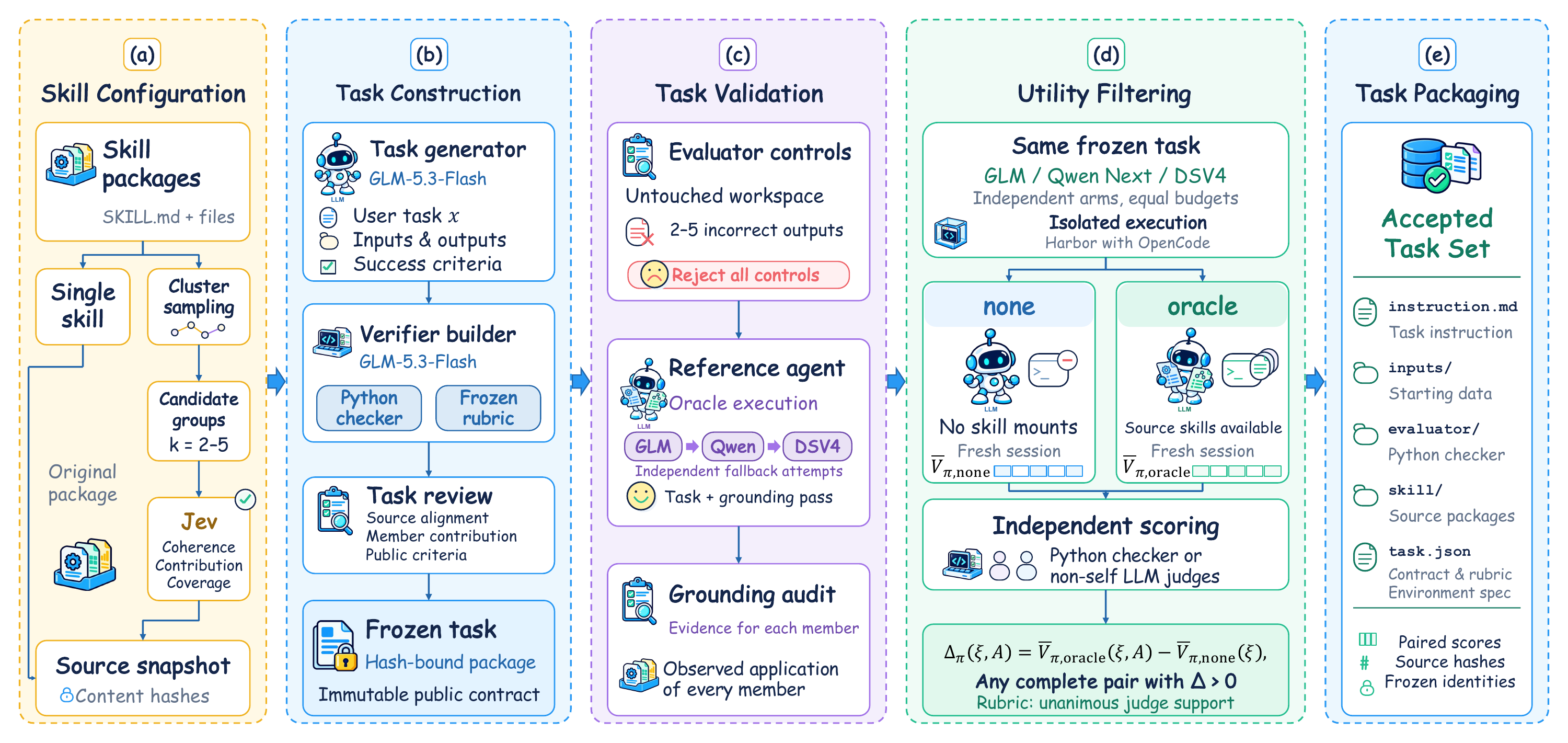}
  \caption{Overview of the skill-grounded data synthesis framework. (a) Configure individual skills or construct and review multi-skill groups. (b) Construct and freeze executable tasks and evaluators. (c) Validate evaluator behavior, task solvability, and skill grounding. (d) Compare matched executions with and without source skills to retain tasks with positive skill utility. (e) Package accepted tasks with evaluation specifications, source skills, and provenance.}
  \label{fig:data-synthesis}
\end{figure}

Our data construction pipeline consists of two stages: skill corpus construction and user task synthesis. The first establishes a unified repository of skills, while the second turns individual skills or compatible skill combinations into executable tasks with corresponding environments and verifiers. We further validate the constructed tasks and filter them based on their solvability and skill utility. Figure~\ref{fig:data-synthesis} provides an overview of the overall user task synthesis pipeline.

\subsubsection{Skill Corpus Construction}
\label{sec:method:corpus}

We collect skills from online skill collections and code repositories, retaining their metadata, instructions, and available supporting resources. To construct a unified skill repository, we consolidate duplicate records across sources by merging their associated information and filter out entries lacking essential skill information. We also preserve source records to maintain traceability from each consolidated skill to its original sources. This process yields a repository $\mathcal{S}$ containing 965K skills. The resulting corpus serves as both the search space for skill search and the source of capabilities for task synthesis. However, the availability of skills alone does not establish whether they can help an agent accomplish a task. We therefore synthesize executable tasks grounded in these skills and evaluate their solvability and utility, as described next.

\subsubsection{User Task Synthesis}
\label{sec:method:synthesis}

Given a source skill set $A\subseteq\mathcal{S}$, we synthesize an executable task $\xi=(x,I,E,V)$. Each source set contains either a single skill or a group of two to five skills. As illustrated in Figure~\ref{fig:data-synthesis}, our synthesis pipeline consists of five stages: skill configuration, task construction, task validation, utility filtering, and task packaging.

\textbf{Skill Configuration.}
As shown in Figure~\ref{fig:data-synthesis}(a), we consider two types of skill configurations: individual skills and multi-skill groups. Individual skills enter the synthesis pipeline directly, whereas multi-skill groups are constructed through cluster-based sampling and assessed by Jev~\citep{almeida2026jev}, instantiated with {StartLux-Decision-v1-35B-A3B}. The reviewer checks whether the group members support a coherent task goal, provide distinct contributions, and contain sufficient source information. Only groups that pass this review proceed to task synthesis. 

\textbf{Task Construction.}
As shown in Figure~\ref{fig:data-synthesis}(b), we use {GLM-5.3-Flash}~\citep{glm5team2026glm5} for task generation, verifier construction, and review, with separate prompts for each role. Given a skill configuration, the generator creates a self-contained task specifying the user instruction $x$, initial inputs $I$, required deliverables, and observable success criteria. For multi-skill tasks, it additionally specifies each skill's intended contribution to the shared goal and supporting source evidence.

The verifier builder then selects an evaluation route based on the success criteria, without changing the task specification. \emph{Verified tasks} have criteria that can be objectively checked through executable behavior in $E$ and are assigned programmatic checkers. Tasks requiring semantic assessment, including those with mixed objective and semantic criteria, are treated as \emph{open tasks} and assigned frozen rubrics aligned with their public success criteria. Each proposal is reviewed for source alignment, requirement clarity, and skill contributions before being materialized in $E$ and frozen with its verifier $V$ and source packages. Private provenance and solution procedures are excluded from public instructions to prevent evaluation leakage. 

\textbf{Task Validation.}
As shown in Figure~\ref{fig:data-synthesis}(c), we validate each task along two dimensions: verifier reliability and task solvability. First, we test whether the verifier correctly rejects incomplete or erroneous outcomes, using the untouched workspace and two to five plausible outputs containing substantive errors as negative controls. Programmatic checkers evaluate verified tasks, while {GLM-5.3-Flash}, {Qwen3.8-Flash-Next}~\citep{qwen2026design}, and {DeepSeek-V4-Flash-0731}~\citep{deepseekai2026deepseekv4} independently assess open tasks against their frozen rubrics. All controls must fail to achieve full completion; for open tasks, all three judges must agree.

We then evaluate task solvability using the same three models as reference agents, each attempting the frozen task in a fresh session with access to all source skills. For open tasks, the two models other than the executing agent evaluate its attempt to reduce self-evaluation bias. An attempt qualifies only if it satisfies all success criteria and passes a grounding audit by {GLM-5.3-Flash}, which verifies that the execution trajectory and resulting artifacts demonstrate source-supported skill application. Multi-skill tasks additionally require evidence of applying every source skill. We allow up to three attempts per task and retain only those with at least one qualifying attempt for subsequent evaluation.

\textbf{Utility Filtering.}
Task solvability alone does not establish that a task is useful for skill search, as an agent may succeed without accessing the source skills. We therefore conduct a final utility check to determine whether the source skills improve task completion. To keep synthesis efficient, the preceding reference attempts use a lightweight native execution loop, while only tasks that pass verifier validation and yield a source-grounded reference solution proceed to this more resource-intensive evaluation. As shown in Figure~\ref{fig:data-synthesis}(d), we conduct this comparison using Harbor with OpenCode~\citep{anomaly2026opencode} as the agent harness. For a fixed agent $\pi$, we independently execute each frozen task under matched \emph{none} and \emph{oracle} conditions, using the same environment, tools, and execution budget. Both conditions start from fresh sessions and workspaces. The none condition provides no skills, whereas the oracle condition makes all source skills in $A$ available to the agent. These executions are independent of the earlier reference attempts, and neither condition receives reference solutions or private evaluation materials.

We evaluate the resulting outcomes using programmatic checkers or rubric judges other than the executing model. Each score is the fraction of satisfied criteria, averaged across judges when applicable. The benefit of providing source skills is defined as
\begin{equation}
  \Delta_{\pi}(\xi,A)
  = \overline{V}_{\pi,\mathrm{oracle}}(\xi,A)
  - \overline{V}_{\pi,\mathrm{none}}(\xi),
  \label{eq:synthesis-skill-benefit}
\end{equation}
where each term is the mean score across configured executions under the corresponding condition.

We retain a task if at least one paired comparison yields positive skill benefit. For rubric-based evaluation, we additionally require positive benefit from every judge and agreement on completion judgments. For multi-skill tasks, $\Delta_{\pi}$ measures the utility of the complete skill group rather than the individual contribution of each member.

\textbf{Task Packaging.}
Tasks that pass utility filtering are packaged into the final collection of nearly 100,000 executable tasks. As shown in Figure~\ref{fig:data-synthesis}(e), each task package contains a public task instruction, starting inputs, source packages, and a private evaluation specification. To support reproducible execution and evaluation, we store programmatic checkers separately and retain the execution environment, expected outputs, and rubric criteria, when applicable, in the task contract.
We retain paired scores, source hashes, and frozen evaluator identities as provenance, linking each task to its source capabilities and measured utility. These records support reproducible search and evaluation, and may also facilitate future agent training.

We further curate \taskbenchmark{}, a 250-task evaluation benchmark, from our synthesized task collections through additional task review and verifier refinement, as detailed in Section~\ref{sec:venusbench}.

\subsection{Utility-Based Training Signal Mining}
\label{sec:method:training}

\begin{figure}[tbp]
  \centering
  \includegraphics[width=\linewidth]{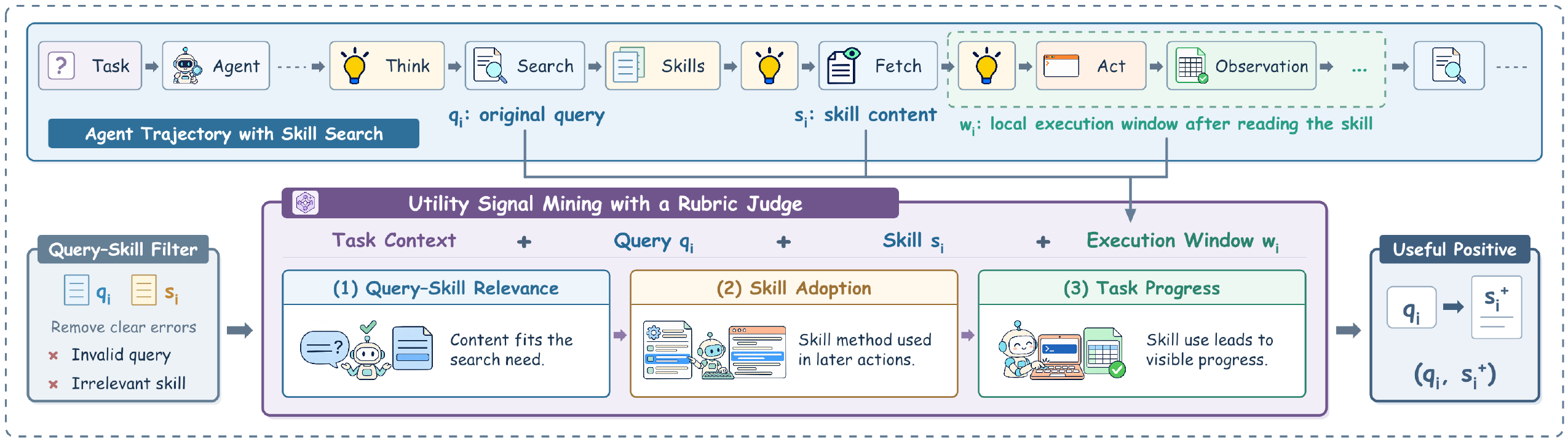}
  \caption{Utility-based training signal mining from agent trajectories. A rubric judge evaluates query--skill relevance, skill adoption, and task progress using the task context, original query, skill content, and post-fetch execution window.}
  \label{fig:utility-signal-mining}
\end{figure}

A skill may appear relevant to a query yet target a different workflow or lack details needed for the task. An agent-oriented skill search model should find skills that are relevant to the query and help the agent make progress on the task. We therefore consider \textbf{utility}: whether retrieved skills help accomplish the downstream task~\citep{qu2025upliftrag,zhang2025utilityannotation}. Inspired by work on agentic search~\citep{zhou2026lrat,liu2026agenticr}, we mine training queries and utility supervision from agent execution trajectories, which record the agent's queries, skill use, and task progress. Figure~\ref{fig:utility-signal-mining} illustrates the mining process.

\textbf{Trajectory Collection.}
We use the tasks constructed in Section~\ref{sec:method:data} to collect agent execution trajectories. Each agent receives task instructions, initial inputs, an execution environment, and a meta-skill interface for searching for and fetching skills. We collect trajectories in OpenCode~\citep{anomaly2026opencode} using four models (Qwen3.8-27B, Qwen3.8-Flash-Next, GLM-5.3-Flash, and DeepSeek-V4.1-Flash) and two retrievers (BM25 and Qwen3-Embedding-0.6B). We record the full execution trajectory of each run, with approximately 392,000 trajectories entering the utility annotation pipeline.

\textbf{Query--Skill Pair Extraction.}
From these trajectories, we extract query--skill pairs $(q_i,s_i)$ by matching each fetched skill $s_i$ to the search query $q_i$ used to retrieve it. For each pair, we extract a local execution window $w_i$ after the skill content is returned. The window contains at most ten agent steps and stops before the next search or when the trajectory ends. We combine this window with the task instructions, query, and skill content as evidence for utility annotation. This evidence connects the agent's search need with its actions after reading the skill.

\textbf{Utility Annotation.}
We first filter the extracted query--skill pairs, removing those with clearly invalid queries or plainly mismatched skills. For the remaining pairs, a rubric judge receives the task context, query $q_i$, visible skill content $s_i$ (with available pre-fetch context and agent reasoning), and local execution window $w_i$, and assesses three criteria in a single annotation call. \emph{Query--skill relevance} checks whether the skill provides applicable knowledge or procedures for the current search need. Skills that address part of the search need also count as relevant. \emph{Skill adoption} requires a concrete correspondence between the target skill's content and subsequent actions, such as applying a rule, command, or method. Reading the skill or expressing an intention to use it is insufficient. \emph{Task progress} requires visible task-relevant progress following that application, such as resolving an error, passing a check, or producing a required artifact. 
This judgment uses the observed local consequence, without requiring final task completion or consulting the final task reward. This process yields approximately 146K queries and 156K positive query--skill pairs.

\subsection{Skill Search Model Training}
\label{sec:method:retriever-training}

\textbf{Training Data.}
We fine-tune Qwen3-Embedding-0.6B on the utility-labeled query--skill pairs from Section~\ref{sec:method:training}. To also support skill search for complete user requests, we include the SkillRet training set of 63,259 queries~\citep{kang2026skillret}. We draw batches from the two sources in a 1:1 ratio, with each batch containing examples from a single source. For agent queries, each training example uses one positive skill, seven hard negatives mined with BM25 and Qwen3-Embedding-0.6B, and in-batch negatives. For SkillRet, each example uses one positive skill and only in-batch negatives.

\textbf{Training Objective.}
When training on a query--positive skill pair, a skill that is positive for that query can also serve as a positive or a negative for another query in the same batch. Standard InfoNCE with in-batch negatives~\citep{oord2018cpc} can then treat that skill as a negative for the original query, pushing it away from the query in the embedding space. This conflict is particularly pronounced for queries with multiple positives, especially in SkillRet. We therefore use a masked InfoNCE objective that excludes known positives from the negative pool.

For a query $q_i$, let $\mathcal{P}_i=\{s_{i,1}^{+},\ldots,s_{i,m_i}^{+}\}$ be its set of known positive skills. When training on a pair $(q_i,s_i^+)$, we keep $s_i^+\in\mathcal{P}_i$ as the target and remove all skills in $\mathcal{P}_i$ from the negative pool $\mathcal{N}_i$, which includes in-batch negatives and hard negatives. Let $z_i(s)$ denote the temperature-scaled cosine similarity between $q_i$ and skill $s$. We average the loss over the $B$ examples in the batch:
\begin{equation}
  \mathcal{L}
  = -\frac{1}{B}\sum_{i=1}^{B}
      \log \frac{\exp(z_i(s_i^+))}
      {\exp(z_i(s_i^+))+\sum_{s\in\mathcal{N}_i\setminus\mathcal{P}_i}\exp(z_i(s))}.
  \label{eq:masked-infonce}
\end{equation}

\section{Benchmarks}
\label{sec:venusbench}

\begin{figure}[!htbp]
  \centering
  \includegraphics[width=\linewidth]{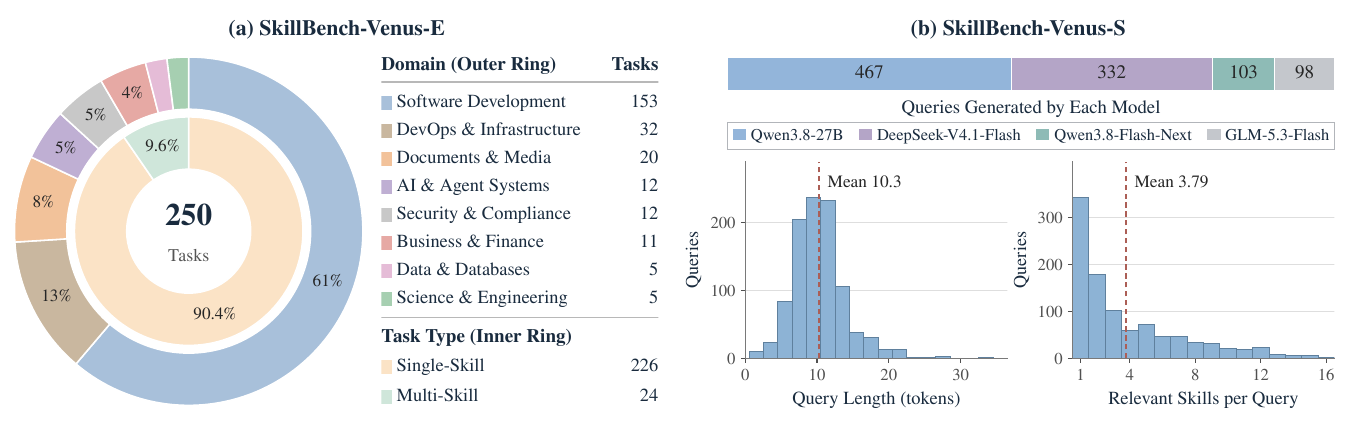}
  \caption{Benchmark composition. (a) Task domains and types in \taskbenchmark{}. (b) Query distributions by model, length, and number of relevant skills in \retrievalbenchmark{}.}
  \label{fig:benchmark-distributions}
\end{figure}

Few benchmarks evaluate whether agents can search for skills as needs arise during execution and use what they find to complete a task. Existing benchmarks typically give agents a set of skills selected for the task or retrieved using the initial task request~\citep{li2026skillsbench,han2026skilluse,su2026sra}. 
We build \taskbenchmarkfull{} to evaluate search and skill use together on executable tasks where access to useful skills can improve performance (Section~\ref{sec:venusbench:e2e}). It also lets us compare how well different skill search models help agents find the skills they need. 
To evaluate skill search models more directly and at lower cost, we build \retrievalbenchmarkfull{}, which measures how well they rank relevant skills for a fixed set of queries collected from agent trajectories (Section~\ref{sec:venusbench:retrieval}).
Figure~\ref{fig:benchmark-distributions} summarizes the composition of the two benchmarks, and Appendix~\ref{app:benchmark-comparison} compares them with existing benchmarks for skill search and skill use.

\subsection{\taskbenchmark{}}
\label{sec:venusbench:e2e}
\begin{figure}[t]
  \centering
  \includegraphics[width=\linewidth]{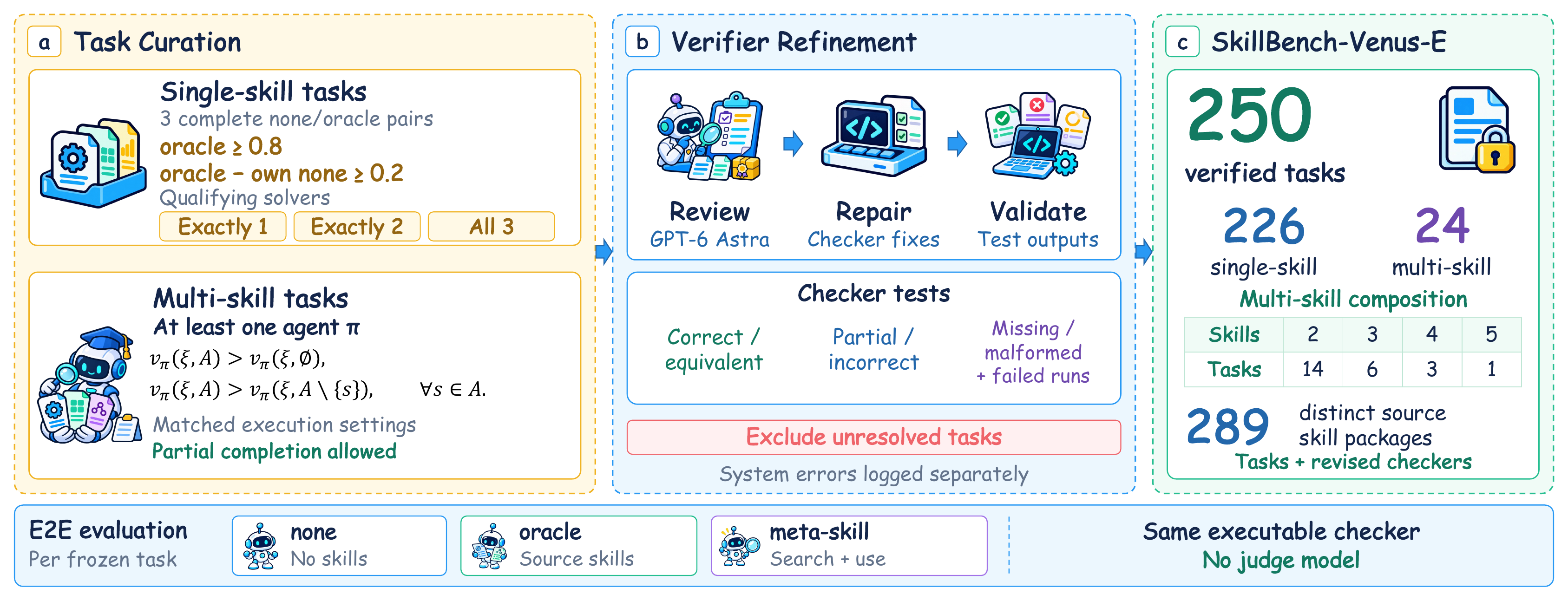}
  \captionsetup{justification=raggedright,singlelinecheck=false}
  \caption{Construction of \taskbenchmark{}. (a) Curate single-skill tasks by solver-specific utility and multi-skill tasks through leave-one-out comparisons. (b) Review tasks, refine executable checkers, and validate scoring behavior. (c) Freeze 250 verified tasks for matched evaluation under none, oracle, and meta-skill access with the same task-specific checker.}
  \label{fig:venusbench-construction}
\end{figure}

We introduce \taskbenchmark{} to evaluate whether LLM agents can effectively discover and apply skills to complete executable tasks. A key challenge in designing such a benchmark is distinguishing genuine skill utilization from task completion driven by agents' intrinsic capabilities. Existing benchmarks, such as SkillsBench~\citep{li2026skillsbench} and SkillEval-v1~\citep{tan2026skt}, exhibit relatively small performance gaps between settings with and without skill access, making this distinction difficult. We address this challenge by curating tasks with demonstrable skill utility, aiming to make the contribution of skill access more distinguishable from agents' intrinsic capabilities. Our benchmark contains 250 verified tasks, substantially more than SkillsBench (87) and SkillEval-v1 (100), enabling more robust comparisons. As shown in Figure~\ref{fig:venusbench-construction}, the construction pipeline consists of task curation and verifier refinement, followed by freezing the resulting executable tasks.

\textbf{Task Curation.}
As shown in Figure~\ref{fig:venusbench-construction}(a), we curate tasks based on their measured skill utility while preserving diversity in solver performance. For single-skill candidates, we consider complete none/oracle comparisons from the three synthesis solvers. A solver qualifies if its oracle score is at least $0.8$ and exceeds its none score by at least $0.2$. We stratify the candidates according to whether exactly one, two, or all three solvers qualify, thereby covering tasks with varying consistency of skill benefits. 
For multi-skill candidates, we use leave-one-out comparisons to assess the contribution of individual source skills. Let $v_{\pi}(\xi,B)$ denote the verifier score of agent $\pi$ on task $\xi$ when provided with skill set $B$. A candidate is retained only if at least one agent satisfies
\begin{equation}
  \begin{aligned}
    v_{\pi}(\xi,A) &> v_{\pi}(\xi,\varnothing), \\
    v_{\pi}(\xi,A) &> v_{\pi}(\xi,A\setminus\{s\}),
      \qquad \forall s\in A.
  \end{aligned}
  \label{eq:venusbench-member-benefit}
\end{equation}
All comparisons use matched task inputs, tools, environments, and execution budgets, and the same agent must satisfy every inequality. This criterion ensures that the complete skill set improves performance over both the no-skill condition and every leave-one-out variant, providing evidence that each source skill contributes to the observed benefit.

\textbf{Verifier Refinement.}
As shown in Figure~\ref{fig:venusbench-construction}(b), we use {GPT-6 Astra} to review candidate tasks and refine their verifiers. The review covers task instructions, initial inputs, source skills, success criteria, checker implementations, and available execution evidence. We exclude tasks with unresolved input deficiencies, ambiguous requirements, or unverifiable objectives, and ensure that each scoring criterion is grounded in the public requirements and supported by the task inputs or source skills.
We further repair missing checks, overly restrictive matching rules, and brittle assumptions about output formats or execution evidence. We test the revised checkers against correct solutions, equivalent implementations, partially correct and incorrect outputs, as well as malformed or missing artifacts. We separately examine evaluator and infrastructure failures to ensure that they are not conflated with task-level failures. These tests help identify both false acceptance and false rejection. Evaluator or infrastructure failures are recorded separately from task scores, and the same revised checker is applied across conditions when reassessing historical executions. Finally, we freeze the verified task packages and their executable checkers for benchmark evaluation.
The final \taskbenchmark{} test set contains 250 verified tasks, including 226 single-skill and 24 multi-skill tasks.

\begin{table}[!tbp]
  \centering
  \captionsetup{justification=raggedright,singlelinecheck=false}
  \caption{Skill access on \taskbenchmark{}, averaged over three runs per task. Partial credit is the mean checker score and full completion is the share of runs earning full credit, both expressed as percentages. Skill use is the percentage of oracle runs that accessed a source skill. $\Delta=S_{\mathrm{oracle}}-S_{\mathrm{none}}$ is the gain in percentage points and $g=100\Delta/S_{\mathrm{none}}$ is the relative gain. The mean row computes $g$ from the mean scores. Bold marks the highest None and Oracle scores.}
  \label{tab:venusbench-validation}
  \begingroup
  \fontsize{10}{12}\selectfont
  \setlength{\tabcolsep}{3pt}
  \renewcommand{\arraystretch}{1.18}
  \resizebox{0.95\linewidth}{!}{%
  \begin{tabular}{@{}clrrrrrrrrr@{}}
    \toprule
    \rowcolor{gtechbg}
    \multicolumn{2}{l}{\textbf{Model}} & \multicolumn{4}{c}{\textbf{Partial Credit}} & \multicolumn{4}{c}{\textbf{Full Completion (0/1)}} & \textbf{Skill Use} \\
    \cmidrule(lr){3-6}\cmidrule(lr){7-10}
     & & None & Oracle & $\Delta$ (pp) & $g$ (\%) & None & Oracle & $\Delta$ (pp) & $g$ (\%) & (\%) \\
    \midrule
    \multirow{6}{*}{\includegraphics[width=1.35em,height=1.35em,keepaspectratio]{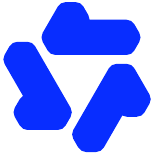}} & Qwen3.5-4B & 11.81 & 34.87 & +23.06 & +195.20 & 1.60 & 22.93 & +21.33 & +1333.33 & 57.73 \\
     & Qwen3.5-9B & 14.17 & 37.59 & +23.42 & +165.20 & 2.27 & 24.53 & +22.27 & +982.35 & 50.27 \\
     & Qwen3.5-27B & 19.85 & 61.69 & +41.84 & +210.72 & 2.93 & 44.93 & +42.00 & +1431.82 & 78.00 \\
     & Qwen3.8-27B & 32.70 & 82.50 & +49.81 & +152.33 & 15.07 & 68.53 & +53.47 & +354.87 & 96.13 \\
     & Qwen3.8-Flash-Next & 34.62 & 83.01 & +48.39 & +139.77 & 16.27 & 69.07 & +52.80 & +324.59 & 99.07 \\
     & Qwen3.8-Max & 37.06 & 83.32 & +46.26 & +124.83 & 19.60 & 71.07 & +51.47 & +262.59 & 98.53 \\
    \midrule
    \raisebox{-.2em}{\includegraphics[width=1.35em,height=1.35em,keepaspectratio]{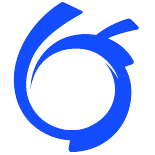}} & Ling-3.1-Flash-1019 & 29.43 & 84.13 & +54.70 & +185.84 & 13.07 & 70.13 & +57.07 & +436.73 & 98.40 \\
    \midrule
    \multirow{2}{*}{\includegraphics[width=1.35em,height=1.35em,keepaspectratio]{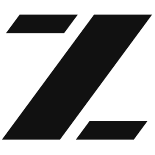}} & GLM-5.3-Flash & 31.75 & 78.21 & +46.46 & +146.35 & 14.93 & 64.13 & +49.20 & +329.46 & 93.33 \\
     & GLM-5.3 & 35.25 & 84.59 & +49.34 & +139.97 & 17.07 & 70.53 & +53.47 & +313.28 & 96.67 \\
    \midrule
    \raisebox{-.2em}{\includegraphics[width=1.35em,height=1.35em,keepaspectratio]{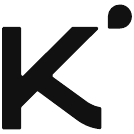}} & Kimi-K3.1-Pro & 38.63 & 81.32 & +42.70 & +110.54 & 20.80 & 70.13 & +49.33 & +237.18 & 95.87 \\
    \midrule
    \raisebox{-.2em}{\includegraphics[width=1.35em,height=1.35em,keepaspectratio]{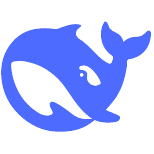}} & DeepSeek-V4.1-Flash-1017 & \textbf{43.57} & \textbf{85.61} & +42.04 & +96.50 & \textbf{28.93} & \textbf{72.67} & +43.73 & +151.15 & 97.07 \\
    \midrule
    \rowcolor{gtechbg}
    \multicolumn{2}{l}{Mean} & 29.89 & 72.44 & +42.55 & +142.32 & 13.87 & 58.97 & +45.10 & +325.26 & 87.37 \\
    \bottomrule
  \end{tabular}%
  }
  \endgroup
\end{table}

\textbf{Quality Validation.}
We compare \emph{none} and \emph{oracle} settings to examine whether providing source skills improves task execution. All 11 models achieve higher scores under oracle access on both metrics (Table~\ref{tab:venusbench-validation}). Mean partial credit rises from $29.89\%$ to $72.44\%$ ($+42.55$ percentage points), while full completion increases from $13.87\%$ to $58.97\%$ ($+45.10$ points). These substantial gains confirm that the curated tasks genuinely benefit from source skills. Nevertheless, skill access does not guarantee task completion: even among the seven models with skill-use rates above $95\%$, full-completion scores range from $68.53\%$ to $72.67\%$. This gap highlights the challenge of effectively applying acquired skills. Moreover, the standard deviation of oracle partial-credit scores across three repetitions ranges from 0.48 to 1.68 percentage points, indicating stable evaluation outcomes. Appendix~\ref{app:venusbench:setup} provides further details on the evaluation protocol and implementation details.

\subsection{\retrievalbenchmark{}}
\label{sec:venusbench:retrieval}

\begin{figure}[!tbp]
  \centering
  \includegraphics[width=\linewidth]{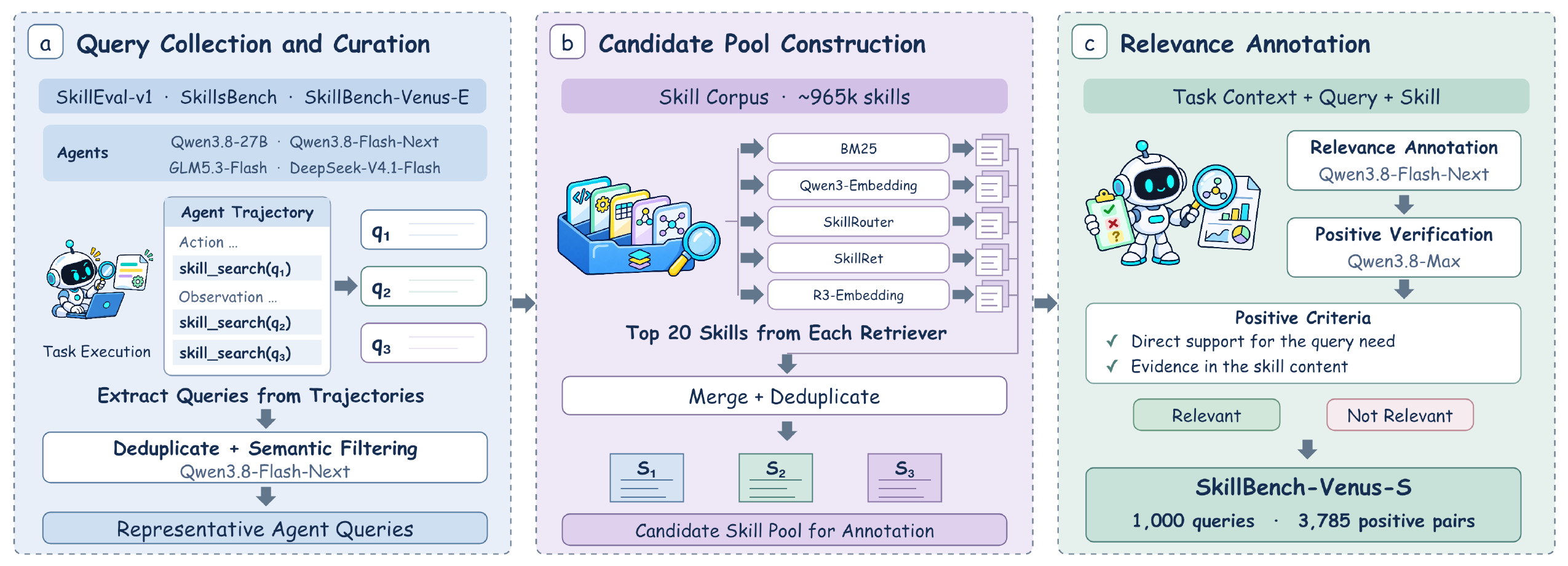}
  \caption{Construction of \retrievalbenchmark{}. (a) Extract and curate search queries from agent execution trajectories. (b) Pool and deduplicate results from five retrievers. (c) Annotate query--skill relevance and verify positive labels.}
  \label{fig:venus-test-construction}
\end{figure}

End-to-end task scores reflect the combined effects of search quality and the agent's use of the retrieved skills, and obtaining these scores requires running the agent with each skill search model. To evaluate the skill search model more directly, we build \retrievalbenchmarkfull{} from search queries that agents issue during task execution. We give each skill search model the same queries and measure how well it ranks skills that address the agent's search need. The resulting offline scores also provide a low-cost proxy for how well a skill search model can support downstream tasks, making it practical to compare skill search model variants before running end-to-end evaluations. Figure~\ref{fig:venus-test-construction} summarizes the construction process.

\textbf{Query Collection and Curation.}
We run agents on tasks from SkillEval-v1, SkillsBench, and \taskbenchmark{}. The agents use four models (Qwen3.8-27B, Qwen3.8-Flash-Next, GLM-5.3-Flash, and DeepSeek-V4.1-Flash). We extract skill search queries from the resulting trajectories. These queries agents' needs that arise during execution, such as configuring a tool or diagnosing an error. Since agents may repeat or rephrase a search, we use Qwen3.8-Flash-Next to filter invalid queries and group queries that express the same need within a task. From each selected group, we use one representative query in its original wording to retrieve candidate skills.

\textbf{Candidate Pool Construction.}
A common practice when annotating traditional retrieval benchmarks such as TREC is to pool top results from multiple retrieval systems~\citep{zobel1998pooling,voorhees2004trec2003,zhang2025utilityannotation}. We follow this approach to build a candidate pool for each curated query. We collect the top 20 results from the corpus of 965k skills (Section~\ref{sec:method:corpus}) with each of five retrievers (BM25, Qwen3-Embedding-0.6B, SkillRouter-Embedding-0.6B, SkillRet-Embedding-0.6B, and R3-Embedding-0.6B). We then merge the lists and remove duplicates (Figure~\ref{fig:venus-test-construction}(b)).

\textbf{Relevance Annotation.}
We pair each query with the skills in its candidate pool and use Qwen3.8-Flash-Next to judge the relevance of each query--skill pair. Along with the query and skill content, we provide the task context to clarify what the agent needs and which constraints apply. We label a query--skill pair as relevant (positive) if the skill offers concrete guidance, such as an applicable method or implementation detail, that addresses the main search need or a substantial part of it. In a second pass, we use Qwen3.8-Max to review the initial positive labels and verify that they are supported by the skill content. 
The final test set contains 1,000 queries from 352 tasks and 3,785 positive query--skill pairs.

\section{Experiments}

\subsection{Experimental Setup}

\textbf{Benchmarks and Metrics.}
We evaluate two aspects of agent-oriented skill search: skill search quality and end-to-end task performance.
For skill search, we use (1) \textbf{\retrievalbenchmark{}}, which captures intermediate capability needs in agent execution and contains 1,000 queries from 352 tasks, 3,785 positively labeled query--skill pairs, and a candidate corpus of 965k skills; 
and (2) \textbf{SkillRet}~\citep{kang2026skillret}, whose evaluation split contains 4,392 complete user requests that may require multiple skills, evaluated against its official corpus of 6,006 skills. 
Evaluating both benchmarks checks whether the skill search model can handle agents' intermediate needs as well as complete user requests. We report Hit@k (H@k), Recall@k (R@k), and NDCG@k (N@k). For complete user requests, we use the same evaluation metrics as SkillRet. For agent queries, we focus on whether relevant skills appear near the top of the ranking.

For end-to-end evaluation, we use (1) \textbf{\taskbenchmark{}}, which contains 250 held-out executable tasks grounded in source skills and evaluates whether agents can discover and apply useful skills; 
and (2) \textbf{SkillEval-v1}~\citep{tan2026skt}, which contains 100 tasks spanning software development, data analysis, security, and business, with each task involving one to three skills. 
We exclude SkillsBench~\citep{li2026skillsbench} from the main experiments because agents make few skill search calls on this benchmark (see Appendix~\ref{app:skillsbench-results}).
The task score is the fraction of available test points earned in a task execution. \# Search Calls is the average number of searches per task. 
To assess search quality, Hit@3 checks whether each search query retrieves a gold skill in its top three results and averages within each task trajectory, while Recall measures the fraction of the task’s gold skills covered by the union of search results over its execution trajectory.

\textbf{Baselines.}
We compare with BM25~\citep{robertson2009bm25}, a lexical search baseline; Qwen3-Embedding-0.6B, Qwen3-Embedding-4B, and Qwen3-Embedding-8B~\citep{zhang2025qwen3embedding}, which provide general-purpose embedding baselines at three model scales; and four skill-specific embedding search models: SkillRouter-Embedding-0.6B~\citep{zheng2026skillrouter}, SkillRep-Embedding-0.6B~\citep{sun2026skillrep}, SkillRet-Embedding-0.6B~\citep{kang2026skillret}, and R3-Embedding-0.6B~\citep{wang2026r3}.
Our main model, \ourmodel{}-Embedding-0.6B, uses Qwen3-Embedding-0.6B as its initialization. Each embedding model uses its released query instruction, and \ourmodel{} uses the same instruction for both training sources and all evaluation benchmarks. The query templates are provided in Appendix~\ref{sec:retrieval-instructions}. 
For end-to-end evaluation, we additionally report \emph{w/o Skill}, which provides no skill access, and \emph{w/ Gold Skills}, which preinstalls the task's gold skills. The latter provides a reference for task performance when the source skills are supplied.

\textbf{Implementation Details.}
We train \ourmodel{} for one epoch using AdamW with a batch size of 256 and weight decay of 0.01. The learning rate warms up to $2\times10^{-5}$ over the first 10\% of training and then decays linearly.
%
For end-to-end evaluation, all skill search models search the same 965k-skill corpus, which includes the gold skills for all evaluation tasks. 
We evaluate Qwen3.8-27B~\citep{qwen2026qwen38}, GLM-5.3-Flash~\citep{zai2026glm53flash}, and DeepSeek-V4.1-Flash-1017~\citep{deepseek2026v41flash1017} as task-solving agents. 
We use the first two agents for in-domain evaluation and DeepSeek-V4.1-Flash-1017 for out-of-domain evaluation.
All three use OpenCode~\citep{anomaly2026opencode} with web search disabled and a shared meta-skill interface for skill search and fetching (Appendix~\ref{app:skill-search-meta-skill}). 
Through this interface, search returns ranked skill names and descriptions, while fetch returns the full skill and saves the skill and its attachments to a designated local directory. 
All models use temperature 1.0 and top-$p$ 0.95, 
with an execution budget of 1,800 seconds per task. 
We run each task independently three times and report results averaged over the three runs.

\begin{table}[tbp]
  \centering
  \captionsetup{position=top,skip=4pt}
  \footnotesize
  \setlength{\tabcolsep}{1.5pt}
  \renewcommand{\arraystretch}{1.12}
  \begin{minipage}[t]{0.489\linewidth}
    \centering
    \setcaptiontype{table}
    \caption{Skill search performance on \textbf{\retrievalbenchmark{}} using queries issued by agents during task execution.}
    \label{tab:retrieval-main}
    \begin{tabularx}{\linewidth}{l*{5}{>{\centering\arraybackslash}X}}
      \toprule
      \multicolumn{1}{c}{\textbf{Skill Search Model}} & \textbf{H@1} & \textbf{H@3} & \textbf{H@5} & \textbf{R@5} & \textbf{N@5} \\
      \midrule
      BM25 & 38.10 & 47.10 & 50.90 & 29.45 & 34.98 \\
      Qwen3-Embedding-0.6B & 29.80 & 39.50 & 44.20 & 23.11 & 27.48 \\
      Qwen3-Embedding-4B & 29.30 & 39.40 & 44.80 & 23.37 & 27.40 \\
      Qwen3-Embedding-8B & 30.00 & 41.00 & 46.40 & 24.47 & 28.55 \\
      SkillRouter-Embedding-0.6B & 22.60 & 31.30 & 34.90 & 16.63 & 20.36 \\
      SkillRep-Embedding-0.6B & 28.30 & 37.60 & 42.10 & 21.95 & 26.18 \\
      SkillRet-Embedding-0.6B & 30.90 & 41.00 & 45.00 & 24.49 & 28.76 \\
      R3-Embedding-0.6B & 30.40 & 39.40 & 45.00 & 23.84 & 27.97 \\
      \rowcolor{gtechbg}
      \ourmodel{}-Embedding-0.6B & \textbf{42.00} & \textbf{51.50} & \textbf{55.90} & \textbf{33.05} & \textbf{38.50} \\
      \bottomrule
    \end{tabularx}
  \end{minipage}%
  \hfill
  \begin{minipage}[t]{0.489\linewidth}
    \centering
    \setcaptiontype{table}
    \caption{Skill search performance on \textbf{SkillRet} using complete user task requests as queries.}
    \label{tab:retrieval-skillret}
    \begin{tabularx}{\linewidth}{l*{5}{>{\centering\arraybackslash}X}}
      \toprule
      \multicolumn{1}{c}{\textbf{Skill Search Model}} & \textbf{H@5} & \textbf{R@5} & \textbf{R@10} & \textbf{N@5} & \textbf{N@10} \\
      \midrule
      BM25 & 48.70 & 35.96 & 41.84 & 32.53 & 34.65 \\
      Qwen3-Embedding-0.6B & 77.89 & 59.76 & 65.04 & 56.99 & 58.94 \\
      Qwen3-Embedding-4B & 77.62 & 59.79 & 66.31 & 56.41 & 58.86 \\
      Qwen3-Embedding-8B & 79.96 & 61.60 & 67.05 & 59.00 & 61.04 \\
      SkillRouter-Embedding-0.6B & 88.87 & 71.65 & 76.45 & 69.51 & 71.37 \\
      SkillRep-Embedding-0.6B & 88.30 & 70.98 & 76.11 & 68.38 & 70.37 \\
      SkillRet-Embedding-0.6B & 91.69 & 80.38 & 85.89 & 76.98 & 79.15 \\
      R3-Embedding-0.6B & \textbf{94.19} & 83.43 & 88.43 & 80.38 & 82.34 \\
      \rowcolor{gtechbg}
      \ourmodel{}-Embedding-0.6B & 93.53 & \textbf{85.04} & \textbf{90.05} & \textbf{80.89} & \textbf{82.85} \\
      \bottomrule
    \end{tabularx}
  \end{minipage}
\end{table}

\subsection{Skill Search Performance}
\label{sec:exp:retrieval}

Table~\ref{tab:retrieval-main} reports skill search results on agent queries from \retrievalbenchmark{}. BM25 is a strong baseline, outperforming every dense baseline on all five metrics. Agents often express their immediate needs through short, keyword-based queries, which may favor lexical matching. 
The other skill-specific search models show no clear advantage over Qwen3-Embedding-0.6B, possibly because their training queries typically describe complete tasks and differ from the queries agents issue for individual steps. \ourmodel{}-Embedding-0.6B shows a \textbf{clear advantage on agent queries issued during task execution}, leading on all five metrics and improving Hit@1 by 12.2 percentage points over its Qwen3 base model.

Table~\ref{tab:retrieval-skillret} reports skill search results on SkillRet's complete task queries. The ranking differs from that on agent queries, with BM25 now trailing all dense search models by a wide margin. This suggests that lexical matching alone is less effective for full task requests. \ourmodel{}-Embedding-0.6B again achieves the strongest overall search performance, leading on four of the five metrics. These results show that \ourmodel{} delivers substantial gains on agent queries while \textbf{maintaining strong skill search performance on complete task queries}.

\begin{table}[!t]
  \centering
  \caption{End-to-end task performance on \taskbenchmark{} and SkillEval-v1. \emph{w/o Skill} gives the agent no access to skills. \emph{w/ Gold Skills} preinstalls the gold skills for each task. Avg. Score is the mean of the two benchmark scores.}
  \label{tab:e2e-main}
  \fontsize{11.17}{13.7}\selectfont
  \setlength{\tabcolsep}{3.2pt}
  \newcommand{\eTwoEheadfont}{\fontsize{9.93}{11.7}\selectfont}
  \newcommand{\eTwoEcolhead}[1]{{\eTwoEheadfont\textbf{#1}}}
  \newcommand{\eTwoEsearchhead}{{\eTwoEheadfont\bfseries\makebox[0pt][c]{\begin{tabular}[c]{@{}c@{}}\# Search\\[-1pt]Calls\end{tabular}}}}
  \newcommand{\eTwoEagenticon}[1]{\raisebox{-.2em}{\includegraphics[width=1.2em,height=1.2em,keepaspectratio]{figures/model_icons/#1.pdf}}\hspace{.4em}}
  \begin{tabularx}{\linewidth}{l*{2}{
    >{\hsize=.925\hsize\linewidth=\hsize\centering\arraybackslash}X
    *{2}{>{\hsize=1.075\hsize\linewidth=\hsize\centering\arraybackslash}X}
    >{\hsize=.925\hsize\linewidth=\hsize\centering\arraybackslash}X}c}
    \toprule
    \multicolumn{1}{c}{\multirow[c]{2}{*}[-1.5ex]{\textbf{Method}}} & \multicolumn{4}{c}{\textbf{\taskbenchmark{}}} & \multicolumn{4}{c}{\textbf{SkillEval-v1}} & \multirow[c]{2}{*}[-1.5ex]{\textbf{Avg. Score}} \\
    \cmidrule(lr){2-5}\cmidrule(lr){6-9}
    & \eTwoEcolhead{Score} & \eTwoEsearchhead & \eTwoEcolhead{Hit@3} & \eTwoEcolhead{Recall} & \eTwoEcolhead{Score} & \eTwoEsearchhead & \eTwoEcolhead{Hit@3} & \eTwoEcolhead{Recall} & \\
    \specialrule{\lightrulewidth}{0.2ex}{0.2ex}
    \multicolumn{10}{c}{\eTwoEagenticon{qwen}\textbf{Qwen3.8-27B}} \\
    \specialrule{\lightrulewidth}{0.2ex}{0.2ex}
    \rowcolor{black!5}
    w/o Skill & 32.70 & \textemdash & \textemdash & \textemdash & 68.80 & \textemdash & \textemdash & \textemdash & 50.75 \\
    \rowcolor{black!5}
    w/ Gold Skills & 82.50 & \textemdash & \textemdash & \textemdash & 81.54 & \textemdash & \textemdash & \textemdash & 82.02 \\
    BM25 & 70.33 & \textbf{3.15} & 52.30 & \textbf{68.38} & 71.83 & 10.20 & 12.06 & 29.86 & 71.08 \\
    Qwen3-Embedding-0.6B & 62.57 & 7.22 & 34.87 & 53.30 & 73.36 & 11.92 & 7.81 & 23.11 & 67.97 \\
    SkillRouter-Embedding-0.6B & 61.37 & 9.49 & 30.92 & 49.81 & 72.83 & 13.76 & 8.20 & 19.25 & 67.10 \\
    SkillRep-Embedding-0.6B & 64.16 & 7.20 & 36.86 & 54.07 & 72.55 & 12.90 & 7.28 & 20.81 & 68.36 \\
    SkillRet-Embedding-0.6B & 64.70 & 6.26 & 40.29 & 59.25 & 72.68 & 12.44 & 12.11 & 28.11 & 68.69 \\
    R3-Embedding-0.6B & 63.73 & 6.89 & 35.31 & 53.47 & 72.59 & 10.66 & 8.39 & 25.48 & 68.16 \\
    \rowcolor{gtechbg}
    \ourmodel{}-Embedding-0.6B & \textbf{71.54} & 3.39 & \textbf{53.07} & 67.97 & \textbf{73.54} & \textbf{8.91} & \textbf{15.60} & \textbf{33.22} & \textbf{72.54} \\
    \specialrule{\lightrulewidth}{0.2ex}{0.2ex}
    \multicolumn{10}{c}{\eTwoEagenticon{glm}\textbf{GLM-5.3-Flash}} \\
    \specialrule{\lightrulewidth}{0.2ex}{0.2ex}
    \rowcolor{black!5}
    w/o Skill & 31.75 & \textemdash & \textemdash & \textemdash & 62.82 & \textemdash & \textemdash & \textemdash & 47.29 \\
    \rowcolor{black!5}
    w/ Gold Skills & 78.21 & \textemdash & \textemdash & \textemdash & 80.04 & \textemdash & \textemdash & \textemdash & 79.12 \\
    BM25 & 67.48 & 2.35 & 46.62 & \textbf{60.92} & 72.38 & 5.64 & 10.53 & 25.22 & 69.93 \\
    Qwen3-Embedding-0.6B & 62.96 & 4.60 & 33.98 & 49.62 & 72.11 & 6.03 & 9.14 & 17.68 & 67.54 \\
    SkillRouter-Embedding-0.6B & 59.96 & 5.49 & 28.26 & 42.98 & 73.20 & 6.32 & 6.56 & 16.33 & 66.58 \\
    SkillRep-Embedding-0.6B & 58.07 & 4.89 & 34.25 & 48.64 & 72.87 & 6.45 & 7.87 & 19.00 & 65.47 \\
    SkillRet-Embedding-0.6B & 63.76 & 3.99 & 36.44 & 50.55 & 72.85 & 5.61 & 11.04 & 21.53 & 68.30 \\
    R3-Embedding-0.6B & 59.66 & 4.44 & 30.97 & 44.58 & 72.05 & 5.89 & 9.88 & 21.44 & 65.85 \\
    \rowcolor{gtechbg}
    \ourmodel{}-Embedding-0.6B & \textbf{69.56} & \textbf{2.26} & \textbf{47.72} & 59.90 & \textbf{73.68} & \textbf{5.34} & \textbf{15.79} & \textbf{29.11} & \textbf{71.62} \\
    \specialrule{\lightrulewidth}{0.2ex}{0.2ex}
    \multicolumn{10}{c}{\eTwoEagenticon{deepseek}\textbf{DeepSeek-V4.1-Flash-1017}} \\
    \specialrule{\lightrulewidth}{0.2ex}{0.2ex}
    \rowcolor{black!5}
    w/o Skill & 43.57 & \textemdash & \textemdash & \textemdash & 69.30 & \textemdash & \textemdash & \textemdash & 56.43 \\
    \rowcolor{black!5}
    w/ Gold Skills & 85.61 & \textemdash & \textemdash & \textemdash & 80.15 & \textemdash & \textemdash & \textemdash & 82.88 \\
    BM25 & 81.36 & \textbf{5.09} & 49.33 & 75.07 & 74.69 & \textbf{36.58} & 8.33 & 48.17 & 78.02 \\
    Qwen3-Embedding-0.6B & 76.95 & 11.76 & 30.82 & 63.68 & 73.53 & 41.52 & 5.33 & 38.72 & 75.24 \\
    SkillRouter-Embedding-0.6B & 75.32 & 14.13 & 26.58 & 55.96 & 74.97 & 45.01 & 4.16 & 28.11 & 75.14 \\
    SkillRep-Embedding-0.6B & 76.02 & 10.97 & 31.40 & 62.70 & 72.90 & 39.21 & 6.11 & 31.28 & 74.46 \\
    SkillRet-Embedding-0.6B & 78.41 & 10.53 & 37.33 & 66.95 & 74.85 & 45.49 & 7.18 & 44.00 & 76.63 \\
    R3-Embedding-0.6B & 78.20 & 11.34 & 31.98 & 62.82 & 73.93 & 45.08 & 7.61 & 43.17 & 76.07 \\
    \rowcolor{gtechbg}
    \ourmodel{}-Embedding-0.6B & \textbf{83.88} & 7.19 & \textbf{49.41} & \textbf{75.94} & \textbf{75.69} & 38.13 & \textbf{9.73} & \textbf{51.56} & \textbf{79.79} \\
    \bottomrule
  \end{tabularx}
\end{table}

\subsection{End-to-End Task Performance}

Table~\ref{tab:e2e-main} reports task scores and search metrics on \taskbenchmark{} and SkillEval-v1. BM25 remains a strong baseline and outperforms the dense baselines in average task score, while the other skill-specific search models generally offer limited gains in search metrics and end-to-end task scores over Qwen3-Embedding-0.6B. 
These patterns are consistent with the offline skill search results on \retrievalbenchmark{}, supporting \textbf{its use as a proxy for downstream task performance}. 
\ourmodel{}-Embedding-0.6B achieves the highest task score among skill search methods on both benchmarks and performs strongly on Hit@3 and Recall, showing that it helps agents \textbf{find relevant skills and solve tasks more effectively}, even for the out-of-domain agent. Beyond these performance gains, \ourmodel{} substantially reduces search calls, making skill search \textbf{more efficient}.

\begin{figure}[tbp]
  \centering
  \includegraphics[width=\linewidth]{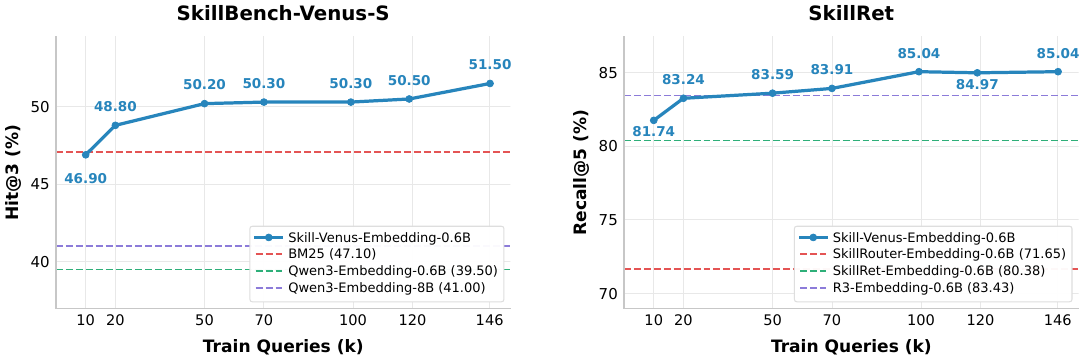}
  \caption{Skill search performance on \retrievalbenchmark{} (left) and SkillRet (right) across training set sizes. Each point reports the result of training \ourmodel{}-Embedding-0.6B for one epoch at the corresponding query count.}
  \label{fig:training-scale}
\end{figure}

\subsection{Further Analysis}
\label{sec:exp:further-analysis}

\textbf{Training Data Size.}
Figure~\ref{fig:training-scale} compares skill search performance on \retrievalbenchmark{} and SkillRet as the number of training queries increases from 10k to 146k. On \retrievalbenchmark{}, only 10k queries yield 46.90 Hit@3, substantially above Qwen3-Embedding-0.6B and Qwen3-Embedding-8B. 
As the number of training queries increases, the results show a clear scaling trend. At 146k queries, the model reaches 51.50 Hit@3, well ahead of all baselines.

On SkillRet, Performance also improves modestly from 81.74 to 85.04. Even the 10k model outperforms SkillRet-Embedding-0.6B and approaches R3-Embedding-0.6B. We hypothesize that two factors contribute to this gain. First, agent queries provide useful training supervision. Second, our positive mask reduces conflicting training signals and helps stabilize training. Our supervision for complete user requests uses the released SkillRet training split, which contains 63,259 queries but only 10,123 skills. Multiple queries can share positive skills; when these queries appear in the same batch, the original in-batch contrastive loss can treat a known positive as a negative~\citep{kang2026skillret}. Our loss masks these known positives, helping use the supervision more effectively.

\begin{figure}[bp]
  \centering
  \includegraphics[width=\linewidth]{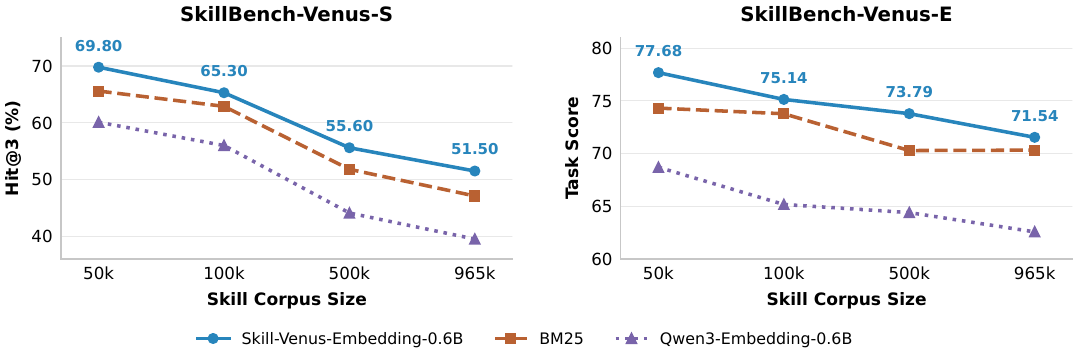}
  \captionsetup{singlelinecheck=true}
  \caption{Skill search and end-to-end task performance across corpus sizes. For the end-to-end evaluation, we use Qwen3.8-27B and average the results over three runs per task.}
  \label{fig:corpus-size}
\end{figure}

\textbf{Skill Corpus Size.}
We compare BM25, Qwen3-Embedding-0.6B, and \ourmodel{}-Embedding-0.6B on corpora of 50k, 100k, 500k, and 965k skills to assess how corpus size affects skill search and end-to-end task performance and whether \ourmodel{} maintains its lead across sizes.
All four corpora retain the gold skills and annotated positive skills from the test sets. We sample from the full corpus to reach the target sizes, keeping each smaller corpus a subset of the larger ones.

\begin{table}[tbp]
  \centering
\captionsetup{position=top,width=\linewidth,justification=justified}
  \caption{Skill search and end-to-end task performance across skill search model sizes. For the end-to-end evaluation, we use Qwen3.8-27B and average the results over three runs per task.}
  \label{tab:model-size-ablation}
  \small
  \setlength{\tabcolsep}{4pt}
  \begin{tabular*}{\linewidth}{@{\extracolsep{\fill}}l*{9}{c}@{}}
    \toprule
    \multicolumn{1}{c}{\multirow[c]{2}{*}[-0.3ex]{\textbf{Skill Search Model}}} & \multicolumn{5}{c}{\textbf{\retrievalbenchmark{}}} & \multicolumn{4}{c}{\textbf{\taskbenchmark{}}} \\
    \cmidrule(lr){2-6}\cmidrule(lr){7-10}
    & \textbf{Hit@1} & \textbf{Hit@3} & \textbf{Hit@5} & \textbf{Recall@5} & \textbf{NDCG@5} & \textbf{Score} & \textbf{\# Search Calls} & \textbf{Hit@3} & \textbf{Recall} \\
    \midrule
    \ourmodel{}-Embedding-0.6B & 42.00 & 51.50 & 55.90 & 33.05 & 38.50 & 71.54 & 3.39 & 53.07 & 67.97 \\
    \ourmodel{}-Embedding-4B & 44.70 & 54.80 & 59.30 & 35.98 & 41.38 & 72.92 & 3.60 & 54.34 & 67.44 \\
    \ourmodel{}-Embedding-8B & \textbf{46.10} & \textbf{56.30} & \textbf{60.00} & \textbf{37.74} & \textbf{43.16} & \textbf{74.02} & \textbf{2.90} & \textbf{56.60} & \textbf{70.15} \\
    \bottomrule
  \end{tabular*}
\end{table}

\textbf{Skill Search Model Size.}
We train Qwen3-Embedding-4B and Qwen3-Embedding-8B on the same data as \ourmodel{}-Embedding-0.6B to study the effect of model size.
Table~\ref{tab:model-size-ablation} compares the three models on offline skill search and end-to-end task performance.
\ourmodel{} shows strong scaling as skill search model size increases, with consistent gains in both offline skill search and end-to-end task scores.
These gains show that our training data and method scale effectively to larger models.

Figure~\ref{fig:corpus-size} compares skill search and end-to-end task performance across corpus sizes.
As the corpus grows, skill search performance declines for all three search models, and end-to-end performance generally follows the same trend.
\ourmodel{} maintains its lead in both evaluations across all four corpus sizes.
To support reproduction and follow-up experiments with different resource budgets, we will release the corpus in multiple sizes.

\enlargethispage{\baselineskip} %
\textbf{Skill Search in an Enterprise Setting.}
To further validate the effectiveness of Skill-Venus in a real-world enterprise setting, we sampled queries from user search logs on Ant ToolCenter, Ant Group's largest internal skill platform, and compared our method with its existing search service.
As shown in Table~\ref{tab:enterprise-skill-search}, Skill-Venus improved Hit@5 by 12.74\% relative to Ant ToolCenter's existing search service. It also ranked relevant skills higher, with relative gains of 5.36\% in MRR@5 and 15.57\% in NDCG@5.
The evaluated collection consists entirely of Ant Group's internal skills and represents an out-of-distribution (OOD) setting for our model.
These gains suggest that Skill-Venus generalizes to Ant ToolCenter's internal skill collection and real user queries.

\begin{table}[t]
  \centering
  \caption{Comparison of Skill-Venus with Ant ToolCenter's existing skill search service on queries sampled from the platform's user search logs. Each value is the percentage improvement over the existing service for the corresponding retrieval metric.}
  \label{tab:enterprise-skill-search}
  \small
  \setlength{\tabcolsep}{4pt}
  \begin{tabular}{@{}lccc@{}}
    \toprule
    Model & Hit@5 & MRR@5 & NDCG@5 \\
    \midrule
    Skill-Venus-Embedding-0.6B & \textbf{+12.74\%} & \textbf{+5.36\%} & \textbf{+15.57\%} \\
    \bottomrule
  \end{tabular}
\end{table}

\section{Related Work}
\label{sec:related}

\subsection{Skill Search}
\label{sec:related:search}

Existing skill search methods broadly follow two approaches. One approach uses graph-based skill search to organize skills through dependency and workflow relations and return skill sets that support execution~\citep{liu2026graphofskills,bai2026skilldag,zeng2026groupofskills}. These methods have mainly been evaluated on small skill libraries containing at most a few thousand skills.
The other approach uses embedding-based skill search to search much larger skill collections, training search models and rerankers to match task queries with skill documents~\citep{zheng2026skillrouter,kang2026skillret}. Building on these models, subsequent studies improve skill search by generating structured pseudo-queries from skill content~\citep{ding2026skill2query}, learning field-specific weights~\citep{goulart2026fieldaware}, and enriching skill documents with capability descriptions~\citep{wang2026capabilitypages}. 
However, the prevailing setup for these skill search models is to select skills before task execution. We argue that skill search should be dynamic, adapting to task progress and emerging capability needs during execution, a direction also explored by SkillBrowser~\citep{ye2026skillbrowser} and JitSkill~\citep{lin2026jitskill} in concurrent work. 
Furthermore, skill search should also be agent-native, with agents deciding when and what to search for based on their own capability needs. \ourmodel{} uses search queries and supervisions extracted from agent trajectories to train a skill search model tailored to these needs.

\subsection{Skill Usage}
\label{sec:related:usage}

Reliable skill use is essential for agents to benefit from external procedural knowledge, motivating extensive work on evaluating and improving this capability~\citep{li2026skillsbench,han2026skilluse,tan2026skt,zhu2026skillcoach,wang2025sage,yang2026skillforge}. 
SkillsBench~\citep{li2026skillsbench} pairs human-authored tasks with expert-curated skills to evaluate agents' task performance. Skill-Use~\citep{han2026skilluse} evaluates whether agents load relevant skills, follow their procedures, and respect their operational boundaries. To improve this capability, SKT~\citep{tan2026skt} synthesizes skill-grounded tasks and verified trajectories for supervised fine-tuning. SkillGym~\citep{wang2026skillgym} automatically builds verifiable environments and collects successful trajectories to train agents for skill use. 
However, these works assume that agents are given a prepared set of skills. In practice, suitable skills must first be retrieved from large collections before they can be used~\citep{su2026sra,liu2026skillusage}. 
Furthermore, agents need to identify their own capability needs during task execution, retrieve suitable skills, and use them effectively to complete the task. \ourmodel{} focuses on this challenge of skill search and use and provides a benchmark to evaluate it.

\section{Conclusion}
In this work, we present \ourmodel{}, an agent-oriented skill search framework that lets agents discover and use skills as needs arise during execution through a reusable meta-skill. We develop a skill-grounded task synthesis pipeline and mine agent trajectories to train the retriever using observed skill adoption and task progress. We introduce \retrievalbenchmark{} and \taskbenchmark{} to evaluate retrieval quality and end-to-end task performance. Extensive experiments show improved retrieval on agent queries, strong performance on complete user requests, and higher task scores with fewer search calls relative to the base retriever. We also publicly release our code, models, training data, benchmarks, and skill corpus to support future research.

\section{Contributions and Acknowledgments}
All contributors of Skill-Venus are listed in alphabetical order by their last names.

\subsection{Core Contributors}
Sunhao Dai$^{\dag}$, Changle Qu, Minghao Tang, Xiaofeng Wu

\subsection{Contributors}
Jingwen Chen,
Shiyu Ni,
Qiao Tong,
Guoqing Wang,
Changshuo Zhang

\subsection{Supervisors}
Keping Bi, Changhua~Meng$^{\dag}$, Weiqiang~Wang

\subsection{Acknowledgments}
We thank the Ant ToolCenter Team for providing Ant Group's internal skills service and for their support during the development of Skill-Venus.

{\let\thefootnote\relax\footnote{$^{\dag}$Corresponding Authors: Sunhao~Dai(sunhaodai@gmail.com), Changhua~Meng(changhua.mch@antgroup.com).}}

\newpage
\bibliographystyle{abbrvnat}
\bibliography{main}

\newpage

\begin{appendix}

\newtcblisting{venusprompt}[1]{
  enhanced,
  breakable,
  listing only,
  listing engine=listings,
  title={#1},
  colback=blue!1,
  colframe=blue!30!black,
  colbacktitle=blue!8,
  coltitle=black,
  fonttitle=\small\bfseries,
  boxrule=0.4pt,
  arc=1mm,
  left=2mm,right=2mm,top=1mm,bottom=1mm,
  before skip=7pt,after skip=9pt,
  listing options={
    basicstyle=\ttfamily\footnotesize,
    columns=fullflexible,
    keepspaces=true,
    breaklines=true,
    breakatwhitespace=true,
    showstringspaces=false,
    tabsize=2,
    aboveskip=0pt,
    belowskip=0pt
  }
}

\section{Retrieval Instructions}
\label{sec:retrieval-instructions}

This appendix lists the query-side instruction templates used by the dense skill search models. The placeholder \texttt{\{query\}} is replaced with the search query. \ourmodel{} uses the same instruction for both training sources and across evaluation benchmarks. BM25 uses the raw query without an instruction prefix.

\tcbset{retrievalprompt/.style={
  colback=blue!1,
  colframe=blue!30!black,
  coltext=black,
  coltitle=black,
  colbacktitle=blue!8,
  fonttitle=\small\bfseries,
  boxrule=0.5pt,
  titlerule=0.4pt,
  arc=1mm,
  breakable
}}

\begin{tcolorbox}[retrievalprompt,title={Qwen3-Embedding-0.6B / Qwen3-Embedding-4B / Qwen3-Embedding-8B}]
\begin{Verbatim}[fontsize=\small,breaklines=true,breakanywhere=true,breaksymbolleft={}]
Instruct: Given a web search query, retrieve relevant passages that answer the query
Query:{query}
\end{Verbatim}
\end{tcolorbox}

\begin{tcolorbox}[retrievalprompt,title={SkillRouter-Embedding-0.6B}]
\begin{Verbatim}[fontsize=\small,breaklines=true,breakanywhere=true,breaksymbolleft={}]
Instruct: Given a task description, retrieve the most relevant skill document that would help an agent complete the task
Query:{query}
\end{Verbatim}
\end{tcolorbox}

\begin{tcolorbox}[retrievalprompt,title={SkillRep-Embedding-0.6B}]
\begin{Verbatim}[fontsize=\small,breaklines=true,breakanywhere=true,breaksymbolleft={}]
Instruct: Given a coding task description, retrieve the most relevant skill document that would help an agent complete the task
Query: {query}
\end{Verbatim}
\end{tcolorbox}

\begin{tcolorbox}[retrievalprompt,title={SkillRet-Embedding-0.6B}]
\begin{Verbatim}[fontsize=\small,breaklines=true,breakanywhere=true,breaksymbolleft={}]
Instruct: Given a skill search query, retrieve relevant skills that match the query
Query: {query}
\end{Verbatim}
\end{tcolorbox}

\begin{tcolorbox}[retrievalprompt,title={R3-Embedding-0.6B}]
\begin{Verbatim}[fontsize=\small,breaklines=true,breakanywhere=true,breaksymbolleft={}]
Instruct: Given a user request, retrieve the agent skill that solves it.
Query: {query}
\end{Verbatim}
\end{tcolorbox}

\begin{tcolorbox}[retrievalprompt,title={\ourmodel{}-Embedding-0.6B / \ourmodel{}-Embedding-4B / \ourmodel{}-Embedding-8B}]
\begin{Verbatim}[fontsize=\small,breaklines=true,breakanywhere=true,breaksymbolleft={}]
Instruct: Given a search query, retrieve skills that provide the knowledge or procedures needed to address it.
Query: {query}
\end{Verbatim}
\end{tcolorbox}

\section{Additional Experimental Setup}
\label{app:venusbench:setup}

\textbf{Implementation Details.} To characterize the benchmark's sensitivity to skill access, we evaluate 11 models from five model families: Qwen~\citep{qwen2026qwen35,qwen2026qwen38,qwen2026design}, GLM~\citep{glm5team2026glm5}, Ling~\citep{inclusionai2026ling31flash}, DeepSeek~\citep{deepseekai2026deepseekv41flash}, and Kimi~\citep{moonshotai2026kimik31pro}. Each model is evaluated three times on all tasks under both conditions. We use Harbor with the OpenCode harness~\citep{anomaly2026opencode}, starting each execution in a fresh session and workspace. Task inputs, tools, sampling parameters, and execution budgets are matched across conditions. In the oracle condition, all source skills are mounted and the agent receives a brief reminder to inspect and use relevant skills; in the none condition, neither is provided. We set the temperature to $1.0$, top-$p$ to $0.95$, with a
context limit of 262,144 tokens and an execution budget of 1,800 seconds per task. We use the highest available reasoning effort for each model, keeping it fixed across both conditions.

\textbf{Evaluation Protocol.} We report two complementary metrics. \emph{Partial credit} is the mean checker score, which rewards partially satisfied task requirements. \emph{Full completion} assigns one only when the original checker score is exactly $1$, and zero otherwise. Both metrics are averaged over all tasks and all three repetitions. All reported comparisons have complete scores. We additionally report the oracle \emph{skill-use rate}, operationalized as the percentage of executions that successfully load a target skill or read its mounted \texttt{SKILL.md}. For multi-skill tasks, accessing at least one target member suffices. This trace-based measure captures skill access rather than successful application.

\begin{table}[!t]
  \centering
  \captionsetup{position=top,singlelinecheck=true}
  \caption{End-to-end results on SkillsBench. Pass Rate measures whether the agent passes all tests for a task.}
  \label{tab:skillsbench-results}
  \small
  \setlength{\tabcolsep}{4pt}
  \begin{tabular*}{\linewidth}{@{\extracolsep{\fill}}l*{8}{c}@{}}
    \toprule
    \multicolumn{1}{c}{\multirow[c]{2}{*}[-0.3ex]{\textbf{Skill Search Model}}} & \multicolumn{4}{c}{\textbf{Qwen3.8-27B}} & \multicolumn{4}{c}{\textbf{DeepSeek-V4.1-Flash-1017}} \\
    \cmidrule(lr){2-5}\cmidrule(lr){6-9}
    & \textbf{Pass Rate} & \textbf{\# Search Calls} & \textbf{Hit@3} & \textbf{Recall} & \textbf{Pass Rate} & \textbf{\# Search Calls} & \textbf{Hit@3} & \textbf{Recall} \\
    \midrule
    BM25 & 43.06 & 1.01 & 25.09 & 23.58 & 63.89 & 1.52 & 25.30 & 24.91 \\
    Qwen3-Embedding-0.6B & 44.91 & 1.04 & 26.55 & 24.49 & 62.96 & 1.71 & 23.99 & 26.81 \\
    SkillRouter-Embedding-0.6B & 47.69 & 1.00 & 22.65 & 16.88 & 62.04 & 1.50 & 17.73 & 19.00 \\
    SkillRep-Embedding-0.6B & 42.59 & 0.91 & 23.77 & 23.08 & 61.57 & 2.56 & 21.73 & 32.99 \\
    SkillRet-Embedding-0.6B & 40.74 & 0.97 & 29.55 & 24.03 & 61.57 & 1.74 & 21.53 & 26.95 \\
    R3-Embedding-0.6B & 46.76 & 1.01 & 35.93 & 28.68 & 62.96 & 1.39 & 25.79 & 26.13 \\
    \ourmodel{}-Embedding-0.6B & 48.15 & 1.00 & 38.90 & 30.71 & 63.43 & 1.55 & 27.52 & 28.70 \\
    \bottomrule
  \end{tabular*}
\end{table}

\begin{table}[!t]
  \centering
  \captionsetup{justification=centering,singlelinecheck=true}
  \caption{Comparison of benchmarks for skill search and skill use.}
  \label{tab:benchmark-comparison}
  \small
  \setlength{\tabcolsep}{3.2pt}
  \renewcommand{\arraystretch}{1.25}
  \begin{tabular*}{\linewidth}{@{\extracolsep{\fill}\hspace{\tabcolsep}}llllrr}
    \toprule
    \textbf{Benchmark} & \textbf{Evaluation} & \textbf{Skill Access} & \textbf{Query Type} & \textbf{Test Set} & \textbf{Corpus} \\
    \midrule
    SkillsBench~\citep{li2026skillsbench} & End-to-end & Task-specific skills & --- & 87 tasks & --- \\
    Skill-Use~\citep{han2026skilluse} & End-to-end + skill use & Task-specific skills & --- & 177 tasks & --- \\
    SkillEval-v1~\citep{tan2026skt} & End-to-end & Task-specific skills & --- & 100 tasks & --- \\
    SRA-Bench~\citep{su2026sra} & Search + end-to-end & Search before execution & Task requests & 5,400 tasks & 26,262 \\
    SkillRouter~\citep{zheng2026skillrouter} & Search + end-to-end & Search before execution & Task requests & 75 queries & \(\sim\)80k \\
    SkillRet~\citep{kang2026skillret} & Skill search & Offline search & Task requests & 4,392 queries & 6,006 \\
    R3-Skill~\citep{wang2026r3} & Skill search & Offline search & Task requests & 5,696 queries & 2,050 \\
    \midrule
    \textbf{\taskbenchmark{}} & End-to-end & \textbf{On-demand search} & \textbf{Agent queries} & 250 tasks & \textbf{965k} \\
    \textbf{\retrievalbenchmark{}} & Skill search & Offline search & \textbf{Agent queries} & 1,000 queries & \textbf{965k} \\
    \bottomrule
  \end{tabular*}
\end{table}

\section{Results on SkillsBench}
\label{app:skillsbench-results}

We evaluate SkillsBench~\citep{li2026skillsbench} with the same skill search interface, corpus, and skill search model configurations as in the main experiments.
We select 72 tasks that can be completed without internet access.
Table~\ref{tab:skillsbench-results} reports results from three runs per task.

Agents make fewer skill search calls per task on SkillsBench than on \taskbenchmark{} and SkillEval-v1, with just one call in about 80\% of SkillsBench trajectories.
Even DeepSeek-V4.1-Flash-1017 averages fewer than two searches per task.
Our trajectory review suggests that SkillsBench task instructions and local materials often give agents enough information to choose a workable approach, after which they proceed to implementation and verification without searching again.
This leaves the skill search model few opportunities to recall the gold skills (2.6 per task on average) and influence task execution.
Scores differ little between skill search models and fluctuate substantially from run to run.
We therefore exclude SkillsBench from the main experiments.
Despite the variation in Pass Rate, \ourmodel{}-Embedding-0.6B outperforms all baselines on Hit@3 and Recall with both agents, showing that its retrieval advantage extends to SkillsBench.

\section{Comparison with Existing Skill Benchmarks}
\label{app:benchmark-comparison}

Table~\ref{tab:benchmark-comparison} compares benchmarks for skill search and use.
In \mbox{\taskbenchmark{}}, agents search a corpus of 965k skills as needs arise during execution.
\mbox{\retrievalbenchmark{}} compares retrievers on the same corpus using a fixed set of queries that agents issued while solving tasks.

\section{Skill Search Meta-Skill}
\label{app:skill-search-meta-skill}

We use the following \texttt{skill-search} meta-skill in our end-to-end evaluations.

\begin{tcolorbox}[retrievalprompt,boxsep=2pt,left=5pt,right=5pt,top=4pt,bottom=4pt,
  title={\mdseries\texttt{SKILL.md}},title after break={\mdseries\texttt{SKILL.md} (continued)}]
\fvset{fontsize={\fontsize{8.4}{9.1}\selectfont},breaklines=true,breakanywhere=true,breaksymbolleft={}}
\begin{Verbatim}
---
name: skill-search
description: Find and load task-relevant skills through the skill-search CLI. Use proactively when the task involves an unfamiliar domain, tool, format, or workflow, or when you are uncertain how to implement or verify a step. Also search for specialized tasks where relevant methods, tools, or conventions could help, even if you already know how to proceed. Skip simple context-only answers, basic text edits, and routine operations that need no specialized guidance.
---

# Skill Search

Discover and load skills that improve the current task. The harness configures the search service and task-local download directory, so call the `skill-search` CLI directly through the shell.

**Avoid oversearching** for a perfect match. Stop searching for the current need once a useful skill is found or query reformulations stop improving results; use what helps or proceed without a skill. For most tasks, aim to stay within **5 search calls**, not to use them all; go beyond this only for clearly distinct unresolved needs.

## When to search

Apply these rules in order:

1. **Reuse adequate guidance first.** If a loaded or previously fetched skill, or explicit task guidance, covers the current capability, reuse it. Search only for what is still missing.
2. **Skip simple work needing no specialized guidance.** Handle straightforward context questions, basic rewriting, and routine code/shell/git operations directly when they need no unfamiliar tool, specialized method, or special format handling. Having all input data in context does not by itself make a task simple.
3. **Search before substantive work where a skill could add value.** Unless the rules above apply, run one focused search for the current need before starting. Familiarity alone is not a reason to skip: a skill may supply methods, tools, or conventions you do not know exist. Also search when an unfamiliar domain, tool, or format, or uncertainty about implementation or verification, creates an uncovered need.

Search for the current stage; reassess as work progresses instead of searching every possible future subtask in advance.

## Workflow

### 1. Describe the current capability needed

Formulate a short, single-intent query for the current stage. Include the domain, operation, and an important constraint when useful. Search for the capability that would help, rather than copying the full task prompt or unrelated output details.

- Split independent needs into separate queries when each becomes relevant.
- Keep tightly coupled operations together when they form one coherent workflow.

Examples:

- Debugging a Python API memory leak → `diagnose Python API memory leak with profiling`
- Reviewing a PostgreSQL migration → `review PostgreSQL migration for locking and rollback safety`
- Testing a Temporal workflow → `test Temporal Python workflows with pytest and time skipping`

### 2. Search and select a useful candidate

```bash
skill-search search --query "<capability needed>" --top-k 5
```

Search returns only each candidate's name, description, and exact `skill_id`, in ranked order. It does not load the skill's instructions or supporting files. Typical output:

```text
{"type":"skill_search","query":"<capability needed>","top_k":5,"result_count":1}

1. <skill name>
   skill_id: <exact-skill-id>
   description: <what the skill helps with>
```

The first JSON line is compact trace metadata: `top_k` is the requested maximum number of results, while `result_count` is the number actually returned and may be smaller. Use the ranked entries below it to choose a candidate; its exact `skill_id` is the value needed by fetch.

Look for a concrete method, tool workflow, or convention that fits the current need; helping with one step is enough. Fetch only promising candidates you intend to use, not the whole result list or a weak match just to use a skill. If none look useful, reformulate when different capability terms or constraints could change the results, or proceed without a skill.

### 3. Fetch the selected skill

Use the exact `skill_id` returned by search, not the candidate's display name, to load the selected skill and its supporting files:

```bash
skill-search fetch --skill-id "<skill_id>"
```

Fetch downloads, validates, and materializes the complete bundle, including `SKILL.md` and any accompanying references, scripts, or assets. The harness configures the task-local destination, normally `/app/.skill-search/skills/<skill_id>/`; use the printed `Directory` as the authoritative location. Typical output:

\end{Verbatim}
\par\noindent\begin{minipage}{\linewidth}
\begin{Verbatim}
```text
{"type":"skill_fetch","skill_id":"<exact-skill-id>","directory":"/app/.skill-search/skills/<exact-skill-id>","cache_hit":false}
Fetched skill: <skill name>
Skill ID: <exact-skill-id>
Directory: /app/.skill-search/skills/<exact-skill-id>
Cache hit: false

<skill_content>
<the complete SKILL.md>
</skill_content>

Associated files:
- /app/.skill-search/skills/<exact-skill-id>/references/example.md
- /app/.skill-search/skills/<exact-skill-id>/scripts/example.py
```
\end{Verbatim}
\end{minipage}\par
\begin{Verbatim}

The first JSON line records the fetch, its materialized directory, and whether an existing local bundle was reused (`cache_hit`). The `skill_content` block contains the complete `SKILL.md` and is the loaded skill: read it and follow the applicable instructions in the current task. `Associated files` lists the local supporting files, or `- none` when there are no attachments; their contents are not included in this list.

Resolve relative paths mentioned by the fetched `SKILL.md` against the printed `Directory`. Use normal file and shell tools to read references, inspect assets, or run included scripts when the skill's workflow calls for them. Read supporting material as needed; the complete bundle remains available in that directory for the rest of the task.

Fetched skills are not registered with the harness's native skill tool. Do not try to load them again through that tool; use the fetch output and local bundle directly.

### 4. Continue and search again when needed

Check that the skill's methods, required tools and dependencies, and input/output assumptions fit the task and environment. Apply the relevant guidance and continue the task within the user's request and permissions. If unsuitable, use another promising candidate or proceed without a skill.

As work progresses, reuse earlier results and local bundles. When a distinct uncovered capability or a materially new constraint emerges, formulate a focused query and repeat the workflow for that need.
\end{Verbatim}
\end{tcolorbox}

\end{appendix}

\end{document}